\documentclass[conference,compsoc]{IEEEtran}

\ifCLASSOPTIONcompsoc
  \usepackage[nocompress]{cite}
\else
  \usepackage{cite}
\fi
\usepackage{url}

\ifCLASSINFOpdf
  \usepackage[pdftex]{graphicx}
\else
\fi

\usepackage{amsmath}

\usepackage{array}
\usepackage{multirow}

\usepackage[caption=false,font=footnotesize,labelfont=footnotesize,textfont=footnotesize]{subfig}

\usepackage{xcolor}

\usepackage{comment}
\usepackage{enumitem}
\usepackage[hidelinks]{hyperref}
\usepackage[capitalise,noabbrev,nameinlink]{cleveref}
\def\showcomments{0}
\ifnum\showcomments=0
    \usepackage{todonotes}
    \newcommand{\fixme}[1]{{\textcolor{red}{[FIXME: #1]}}}
    \newcommand{\checkme}[1]{{\textcolor{orange}{[CHECKME: #1]}}}
    \newcommand{\David}[2][]{\todo[color=blue!30,#1]{D: #2}}
    \newcommand{\Stanley}[2][]{\todo[color=orange!30,#1]{S: #2}}
    \newcommand{\Gururaj}[2][]{\todo[color=magenta!30,#1]{G: #2}}
\else
    \newcommand{\fixme}[1]{}
    \newcommand{\checkme}[1]{}
    \newcommand{\Stanley}[1]{}
    \newcommand{\David}[1]{}
    \newcommand{\Gururaj}[1]{}    
\fi

\begin{document}
\pagestyle{plain}
\bstctlcite{IEEEexample:BSTcontrol}

\title{Time Will Tell: Timing Side Channels via Output Token Count \\ in Large Language Models}

\author{
   \IEEEauthorblockN{Tianchen Zhang \qquad Gururaj Saileshwar \qquad David Lie}
   \IEEEauthorblockA{University of Toronto}
}

\maketitle

\begin{abstract}

This paper demonstrates a new side-channel that enables an adversary to extract sensitive information about inference inputs in large language models (LLMs) based on the number of output tokens in the LLM response. We construct attacks using this side-channel in two common LLM tasks: recovering the target language in machine translation tasks and recovering the output class in classification tasks. In addition, due to the auto-regressive generation mechanism in LLMs, an adversary can recover the output token count reliably using a timing channel, even over the network against a popular closed-source commercial LLM. 

Our experiments show that an adversary can learn the output language in translation tasks with more than 75\% precision across three different models (Tower, M2M100, MBart50). Using this side-channel, we also show the input class in text classification tasks can be leaked out with more than 70\% precision from open-source LLMs like Llama-3.1, Llama-3.2, Gemma2, and production models like GPT-4o.
Finally, we propose tokenizer-, system-, and prompt-based mitigations against the output token count side-channel.

\end{abstract}

\IEEEpeerreviewmaketitle

\section{Introduction}
\label{sec:introduction}
Large language models (LLMs)~\cite{NIPS2017_3f5ee243, touvron2023llama2} are gaining rapid adoption through web applications
such as OpenAI's ChatGPT \cite{openai_introducing_2022}, Microsoft's Copilot \cite{spataro_introducing_2023}, and Amazon's Rufus~\cite{chilimbi_technology_2024}. 
The wide use of LLMs in our daily lives has been driven by their strong performance on tasks such as question-answer, code generation, translation, etc..
However, this also raises concerns about the ability of LLM-based applications to safeguard the privacy of users interacting with them.

Previous work has demonstrated privacy breaches in machine learning systems using attacks such as membership inference~\cite{shi2023detecting,meeus2024did}, training data extraction~\cite{carlini2021extracting}, and prompt injection attacks~\cite{liu2024promptinjection}. More recently, researchers have also started to study privacy breaches via side-channels in LLMs, which leak information by monitoring only the metadata of responses~\cite{299888,song2024earlybirdcatchesleak}. For instance, \cite{299888} demonstrates that the size of the plaintext representation of each token (i.e. the token length) can be used to recover the contents of LLM messages. An LLM implementation that streams responses back token-by-token enables a network adversary to recover such token lengths and statistically guess the content of the responses. Other work on side-channels has exploited performance optimizations in LLMs such as speculative decoding~\cite{wei2024privacy}, and shared semantic caches and KV Caches~\cite{song2024earlybirdcatchesleak}  to leak user prompts based on variations in packet sizes and execution time respectively.

In this paper, we propose the first output token count side-channel in LLMs to leak information during inference. Our key observation is that applications based on LLMs, such as translation or text classification with explanation, can have variations in their output token counts based on private attributes, such as output language in translation or output class in classification. 
While some of these variations have been observed by others, such as biases in tokenizers due to resource-poor languages~\cite{petrov_language_2023, ahia2023languages}; others are newly discovered in this paper, such as natural biases in the explanation for a particular classification. In the case of the previously observed biases in languages, we are the first to take advantage of them to leak sensitive information.

To record the output token count, we propose that an attacker can either rely on the LLM implementation to leak it, by observing the tokens streaming one at a time over a network, or by exploiting a newly proposed timing side-channel, that exploits the autoregressive mechanism that LLMs use to generate output to recover the number of output tokens. The autoregressive mechanism in LLMs generates each output token one at a time, with a constant time per token. This autoregressive generation is an inherent feature of \textit{all} modern transformer-based LLMs~\cite{NIPS2017_3f5ee243,touvron2023llama2}, making them all potentially vulnerable to this side-channel. An attacker can thus learn the number of output tokens by just measuring the time an LLM takes to respond to a query, as the autoregressive loop dominates the LLM response generation time. Moreover, as the length of the input prompt can also affect the length of the LLM output, we find that our attacks can be further improved by taking the input length into account, if it is available to the adversary.

\begin{figure*}[h!]
    \centering
    \includegraphics[width=0.9\linewidth]{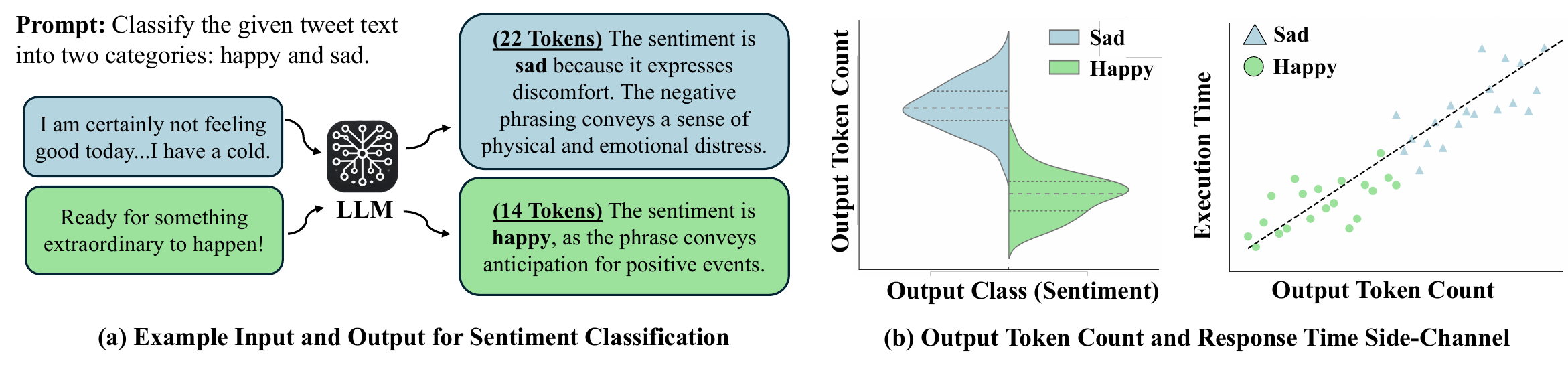}
    \vspace{-0.1in}
    \caption{Example of our side-channel attacks on text sentiment classification with an LLM. (a) For a given text input, the LLM outputs the classified sentiment and also provides an explanation that can vary in length. (b)
    Our token-count side-channel attack leaks the output class (sentiment) based on the bias in the output token counts, or by measuring the execution time, as a proxy for the output token count.}
    \vspace{-0.2in}
    \label{fig:introduction}
\end{figure*}

\smallskip

\noindent \textbf{Effectiveness of our Attacks.}
As LLMs exhibit biases in the token encoding efficiency across languages, using the output token count to compute the output token density (output-bytes/token) and output/input byte ratio, enables our attack to identify over 12 languages across three different multilingual models, Tower, M2M100, MBart50, with an average precision of 82.5\%, 77.2\%, 80.5\% respectively. 
Moreover, languages with distinctive morphology, such as Chinese, Russian, Hindi, Korean, and Arabic, reach almost 100\% precision. 
Similarly, our end-to-end remote attack over the network on AWS-server hosted models, using response times instead of token counts, has high average precision for Tower (83.2\%), M2M100 (75.8\%) and MBart50 (81.3\%). 

We also observe that LLMs exhibit inherent bias in explanation lengths in a range of text classification tasks. For instance, as shown in \cref{fig:introduction}, an LLM performing sentiment analysis and generating explanations may produce longer responses for one class over the other. When profiled at scale, we see that such biases exist in a task-dependent manner across tens of tasks in all LLMs we profile (Llama-3, Gemma2, GPT4). This bias enables an adversary observing output token counts to infer the output class of the classification task.
Moreover, we show that these biases can be exacerbated if there is accompanying bias in the in-context few-shot prompting examples, which amplifies the output token count signal and increases our attack success rates by 15\% on average.
In this scenario, our output token side-channel attack has a success rate of 81.4\% for Gemma2-9B, 72.3\% for Llama3.2-3B, and 86.9\% for GPT-4o. In the end-to-end timing attack, an attack on a Gemma2-9B model achieves a similar success rate of 79.7\%, while a remote timing attack over the internet on production GPT-4o model, hosted on OpenAI servers, has 74.7\% success rate.

\smallskip
\noindent \textbf{Summary of Contributions:}
\begin{enumerate}
    \item \textbf{New Timing Side-Channel:} We identify a new timing side-channel in LLMs due to variations in output token counts. These variations, correlated with task-specific sensitive attributes, can be exploited due to the autoregressive nature of LLMs, enabling attackers to infer private information using execution time measurements.

\item \textbf{Demonstrate Tokenizer Bias is a Privacy Risk:} We show that language-based differences in token density and output token counts can leak language preferences through our side-channel, reframing these biases as a privacy risk rather than solely a fairness issue.

\item \textbf{Highlight Token-Count Bias in Text Classification:} We demonstrate that inherent biases in the token counts of generated explanations across classification tasks can leak sensitive output classes. We show that few-shot prompting can amplify these biases, increasing the success rate of token-count side-channels.

\item \textbf{End-to-End Remote Attacks over the Network:} 
We show remote timing attacks leaking output language in translation tasks (75\% to 83\% success rates) and output class in text classification (with 75\% success rates). 

\item \textbf{Mitigations}: To mitigate these side channels, we propose strategies at the tokenizer, system, and prompt level to prevent these private attributes being leaked.
\end{enumerate}

\section{Background}
\label{sec:background}
\subsection{LLM Autoregressive Decoding}
Autoregressive generation in LLMs generates tokens sequentially,  
with each token depending on all prior tokens.  
For a sequence \( X = (x_1, x_2, \dots, x_t) \),  
the next token \( x_{t+1} \) is predicted using the probability distribution  
\( P(x_{t+1} \mid x_1, x_2, \dots, x_t) \).  
This sequential dependency limits generation to one token per iteration.

LLM token generation has two phases: prefill and decode. 
The prefill stage processes the input prompt and generates the Key-Value (KV)-cache. 
This stage is highly parallelizable, and thus its time, measured by Time To First Token (TTFT), is inconsequential as a proportion of the overall generation time compared to the decode stage.
The decode stage is more memory-bound due to the sequential dependency of newly generated tokens.
While computations per token mildly increase as output length grows, the time to decode each token, measured by Time Per Output Token (TPOT), is relatively constant. 
Overall, the execution time ($T$) to generate a sequence (\(X_n\)) with $n$ tokens scales linearly with the number of output tokens (\(n\)), \textit{i.e.}, $T(X_n) \propto n$.

LLM-based applications operate in either \textit{streaming} or \textit{non-streaming} modes.  
In \textit{streaming} mode, used by interactive applications like chatbots~\cite{openai_2024_gpt4o,spataro_introducing_2023}, tokens are delivered to the user immediately as they are generated.  
This enables adversaries to observe fine-grained TPOT and token counts.  
In \textit{non-streaming} mode, often used for text processing, only the total generation time is observable.  
In \cref{sec:attack_on_translation} and \cref{sec:attack_on_prompt}, we assume streaming mode and demonstrate attacks leveraging token counts.  
In \cref{sec:evaluation}, we show that our attacks remain effective even with the non-streaming mode, using only overall execution time.

\subsection{Tokenization}
\label{sec:tokenization}
Tokenization is the process of splitting text into smaller units, or tokens, for processing by language models. Subword tokenizers, such as Byte Pair Encoding (BPE) \cite{sennrich2016neural}, WordPiece \cite{wu2016google}, and SentencePiece \cite{kudo2018sentencepiece}, have become standard in large language models (LLMs) since the introduction of the transformer architecture \cite{NIPS2017_3f5ee243}. These methods balance a fixed vocabulary size with the ability to handle rare or unseen words by leveraging meaningful subwords that reflect language morphology.

Despite their widespread use, subword tokenizers exhibit biases. 
With BPE tokenizers used in English-centric LLMs (Llama, GPT, Mistral etc.), tasks in resource-poor languages may require up to 13 times more tokens than in English due to over-fragmentation \cite{petrov_language_2023, ahia2023languages}. This bias stems from unbalanced training datasets and suboptimal vocabulary allocation, which disproportionately affect resource-poor languages by reducing compression rates \cite{goldman2024unpacking}, increasing latency, and inflating API costs.
Although tokenizers in multilingual models (e.g., M2M100~\cite{fan_m2m100}, BLOOM~\cite{Bloom}) exhibit less bias, several languages still require token counts that are 2.5$\times$ higher than similar sentences in English.

Recent work proposes methods to address these issues, including improving sampling efficiency \cite{zheng2021allocating}, better vocabulary allocation \cite{lian2024scaffold}, and minimizing over-fragmentation in multilingual settings \cite{liang2023xlmv}. However, language bias remains pervasive. In this paper, we demonstrate that such biases not only lead to performance disparities but also pose privacy risks by revealing a user's language preferences.

\subsection{Few-Shot Prompting}
LLMs are increasingly applied to text classification tasks, such as sentiment analysis \cite{devlin2019bert}, intent detection \cite{chen2019intent}, and topic classification \cite{grootendorst2022bertopicneuraltopicmodeling}, where task-specific data is often required. Traditional methods like fine-tuning demand large labelled datasets and significant compute resources~\cite{howard2018universal}, making them impractical for many applications.

Few-shot prompting~\cite{brown2020fewshot} offers an efficient alternative by providing a small number of task-specific examples in the input prompt. This enables the model to perform the task without expensive fine-tuning, leveraging its ability to learn in context from the examples provided. This approach is useful in scenarios where data is scarce or rapidly evolving and can provide performance comparable to fine-tuning.

Research on few-shot prompting has explored biases in the text classification output and how provided demonstrations have helped improve the performance. For example, prior works~\cite{zhao2021calibrate} identify that answers from the LLM can be biased towards a certain direction. Furthermore, prior works show that including explanations and optimal demonstration examples in prompts can improve accuracy~\cite{ye2022unreliability,nie2022improving,ma2023fairness}.

In contrast, we examine whether text classification tasks have natural biases in output lengths, and whether few-shot prompting can unintentionally enhance signals in output lengths, affecting the side channels leaking output classes.

\section{Threat Model}
\label{sec:threat_model}
\begin{figure}[h!]
    \centering
    \includegraphics[width=\linewidth]{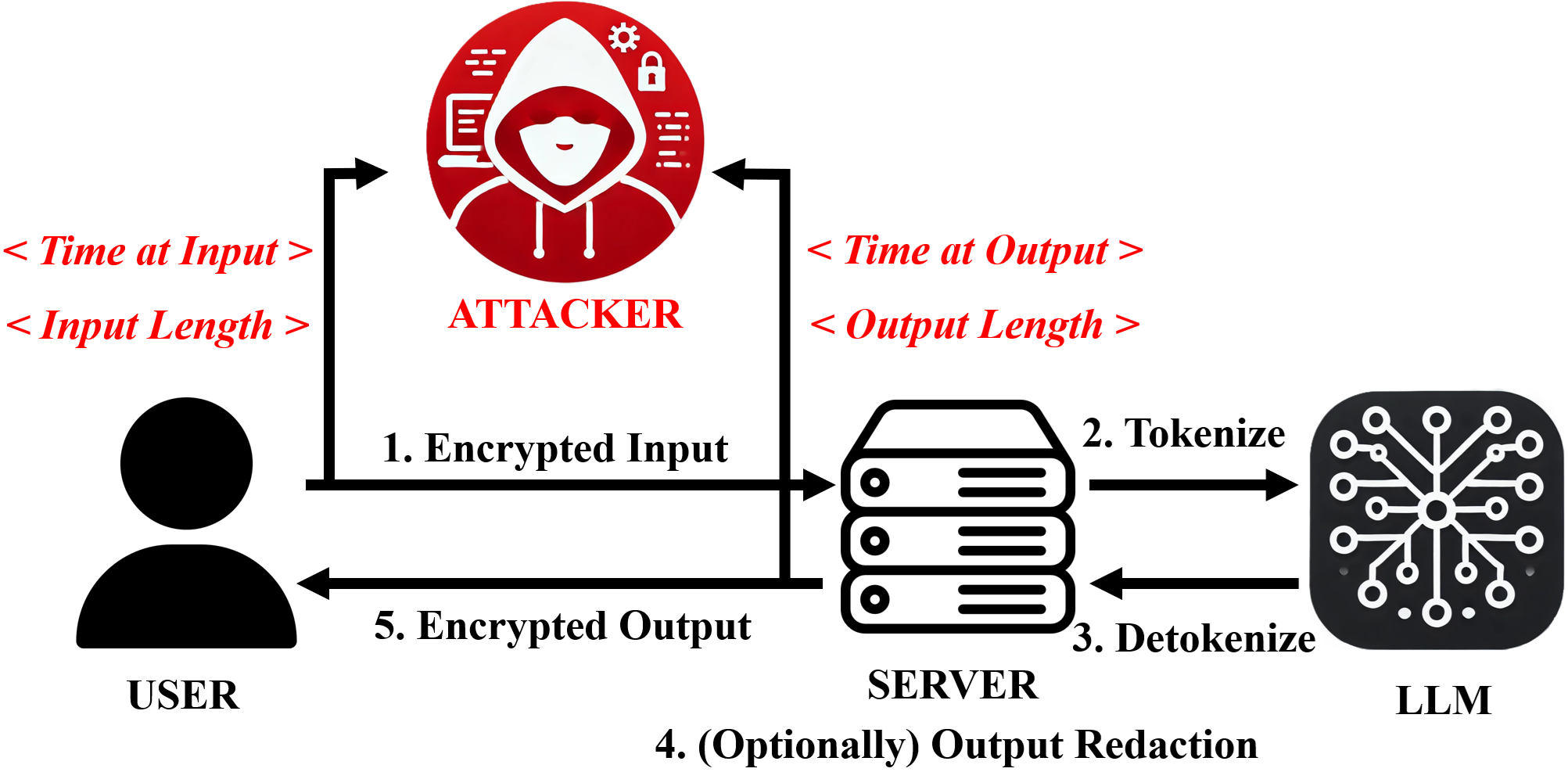}
    \caption{Threat model. We assume the attacker is network-based, and cannot inspect the contents of the encrypted input or output, but can monitor the overall response time, and length of the entire encrypted input or output.}
    \label{fig:threat_model}
\end{figure}

\textbf{Attack Scenario.} As shown in \cref{fig:threat_model}, we study a typical scenario where an LLM-based application is running on the server and the user interacts with it over the internet. The network packets are encrypted, but preserve the plaintext length and are observable by the attacker, similar to prior work \cite{299888}. However, unlike prior work, we do not require the LLM to be in a streaming mode; our attack works even if the LLM operates in non-streaming mode, when the response is delivered all at once, as opposed to token-by token. 

\textbf{Attacker Capabilities.} We assume the attacker knows the task for which the user is using the LLM-based application. 
Additionally, the attacker can also interact with the application as a regular user, allowing the attacker to obtain profiles of how the LLM processes requests. In this scenario, we assume the attacker can obtain three metrics: the total length of the encrypted input text, the total length of the encrypted output text, and the LLM response time, i.e., time from the input packet being sent to the server to the response being received by the user. 
In \cref{sec:attack_on_translation} and \cref{sec:attack_on_prompt}, we initially assume that the attacker can obtain the exact output token count, observable in the case of a streaming LLM, to get an upper bound for the success rate of our attack. 
We then discharge that assumption in \cref{sec:evaluation}, and measure the efficacy of a network attacker who is limited to observing a non-streaming LLM. In this scenario, attacker has to approximate the output token count using the LLM response time.

\textbf{Attacker's Goal.} %
The attacker's goal is to leak private information about the user, such as the target language in the translation task or the output class in the classification tasks. The attacker seeks to leverage the output token count timing side-channel and the length of the entire encrypted output and input, to leak private attributes with high accuracy.

\section{Attack on Translation Workloads}
\label{sec:attack_on_translation}
\subsection{Application Scenario}
In recent years, LLMs are becoming the premier choice for machine translation. Web-based translation services such as Google Cloud's adaptive translation\cite{google_cloud_translate} and DeepL's next generation translator\cite{deepls_nextgen} already deploy LLM-based translation. 
Users of such translation services typically translate a foreign language into their own language, demonstrating their language preference. Language preferences are highly correlated with other private attributes, such as ethnicity and nationality, and Europe's General Data Protection Regulation (GDPR) regulations consider any such information that can identify an individual either directly or indirectly as personally identifiable information, which must be protected\cite{gdpr_article_4}.
Thus, the translation service provider is liable to safeguard a user's language preference and ensure that it remains protected from attackers.

We assume the user sends multiple requests to translate inputs in a fixed source language into a target language.
In this context, the attacker aims to recover the unknown output language of the translation knowing the source language.
With the capability to monitor the LLM request and response timings, and total input and output length, we show that the attacker can deduce the target language the user is translating into with high precision. 
Here, we assume that the attacker has perfect access to the output token count. Later in \cref{sec:evaluation}, we show that with timing information, the output token count can be accurately estimated.

\subsection{Attack Overview}
LLMs employ subword tokenizers typically based on Byte Pair Encoding (BPE), constructing vocabularies influenced by token frequency.  
Biases in training data cause token vocabularies to favour high-resource languages, resulting in more tokens for similar content in lower-resource languages.~\cite{petrov_language_2023, ahia2023languages}
Language morphology differences, such as word length and prefix/suffix usage, further affect vocabulary construction and token densities.  
We leverage this bias to create a side channel where attackers can infer the target language based on token density and input length.  

\begin{figure}[tb!]
  \centering
  \includegraphics[width=\linewidth]{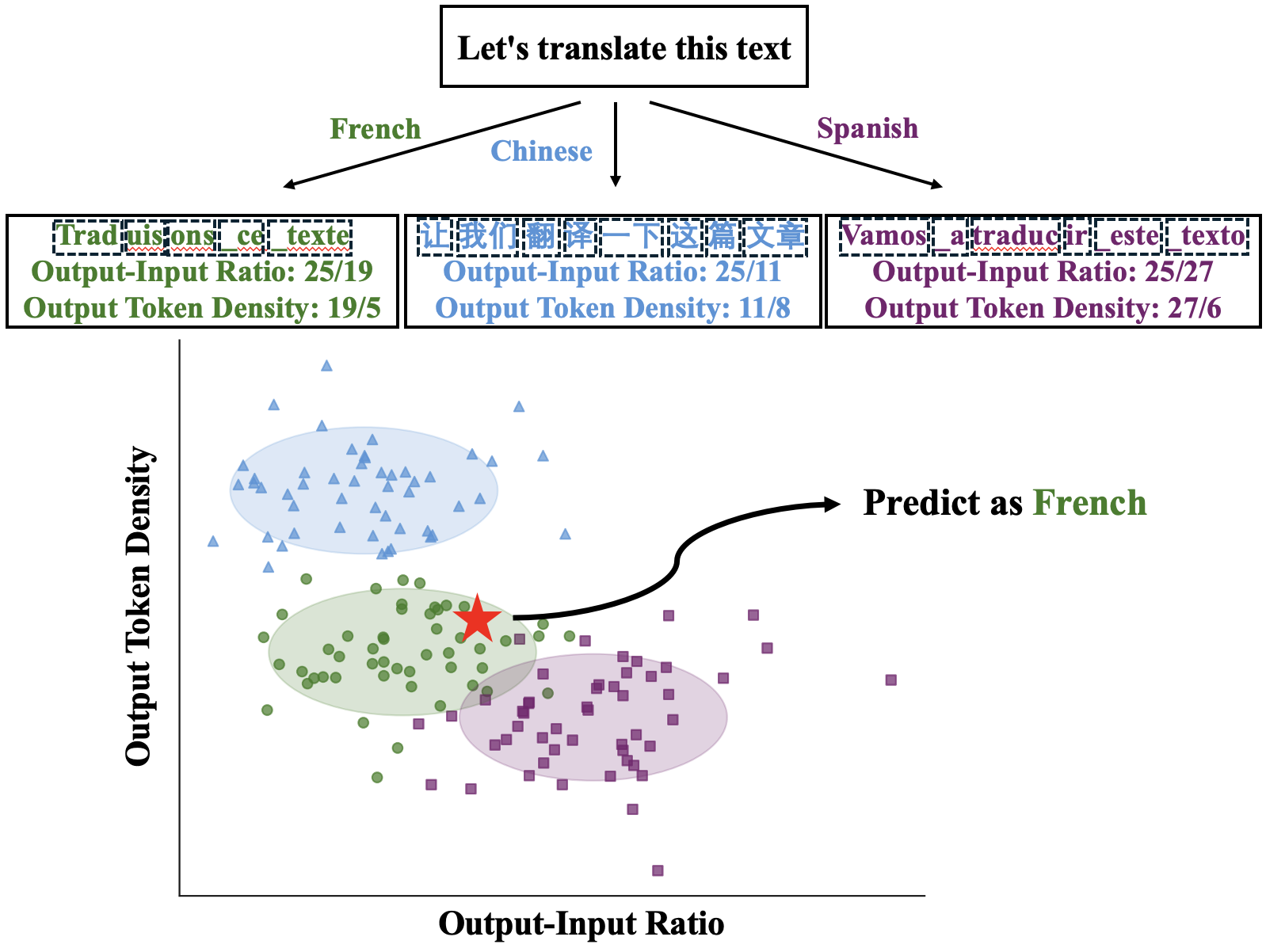}
  \caption{Overview of Attack on Translation. A user translates text in a source language (e.g., English) to any target language; the goal of the attack is to leak this target language. By profiling output token density and output/input bytes ratio for different languages offline, and by observing these for a given translation, the attacker can leak the user's target language.}
  \label{fig:translation}
\end{figure}

\cref{fig:translation} shows an example of an attacker profiling the characteristics of three different languages (French, Chinese, Spanish), by requesting the input ``\textit{Let's translate this text}'' to be translated into each of these languages.
At the same time, the attacker measures the number of output tokens, and lengths of the input request and output response in natural language.
For the same input, the output in each language has a different \textit{token density}\footnote{Note that we use output token density, instead of just \# output tokens, since we seek a metric that is independent of the output length, as the inputs and outputs for a given user request can be arbitrarily long.} (length of output in bytes / \# output tokens), due to the tokenizer's vocabulary differences. 
Moreover, due to the morphological differences, these responses have different \textit{output/input byte ratios}: sentences with the same semantic meaning can have different lengths in bytes in different languages.

The attacker can thus develop a 2D profile for different target languages using these two metrics: (1) \textit{output token density} and (2) \textit{output-input byte ratio}, using a large number of inputs offline.
Subsequently, by monitoring a user's translation request and response, and calculating the output token density and output-input bytes ratio, the attacker can map these to a point in this 2D profile and recover the user's unknown target language with high precision.

\subsection{Attack Implementation}
\noindent \textbf{Profiling Phase.} 
In the profile phase, the attacker sends 1000 requests per language to the LLM to be translated and measures the \textit{Output Token Density} and \textit{Output-Input Ratio} for each request-response, formalized as follows.
For a request and response pair, with input length in bytes as $L_{\text{input}}$, and the output length in bytes as $L_{\text{output}}$,
the (1) \textit{output token density} and (2) \textit{output-input ratio} are calculated as:
\begin{equation}
\textit{Output Token Density} = \frac{L_{\text{output}}}{\textit{\# of Generated Tokens}}
\end{equation}
\begin{equation}
\textit{Output-Input Ratio} = \frac{L_{\text{output}}}{L_{\text{input}}}
\end{equation}

Using the profiling data, the attacker learns two Gaussian distributions for the output-input length ratio and the token density. These two distributions are used to construct a two-dimensional Gaussian Mixture Model (GMM) \cite{dempster_gmm} for each target language (given a model and source language) that describes its characteristics (see \cref{fig:gmm_tower}). 
We use this profiled 2D GMM to launch our attack next.

\smallskip
\noindent \textbf{Attack Phase.}
In the attack phase, we assume the user sends multiple translation requests (1 to 50) to the LLM for the same source-target language pair. 
The attacker gathers the input and output encrypted text lengths and the generated output token count (to calculate token density) across these requests and constructs a new GMM based on this data. 
We use the Bhattacharyya Distance~\cite{3142ae09} to map this GMM to the most similar GMM learned in the profiling phase, to leak the target language. 

\subsection{Experimental Methodology}
\noindent {\bf Models.}
For testing our attack, we prioritize multilingual models that consider language parity~\cite{petrov_language_2023}, and achieve high translation performance for a wide range of languages, measured by BLEU scores. 
Based on these constraints, we test on the following models: M2M100 \cite{fan_m2m100}, MBart50 \cite{tang_mbart50} and Tower \cite{alves_tower}.
M2M100, from Facebook, is specialized for direct many-to-many multilingual translation. It uses a single shared vocabulary across all languages, built with a SentencePiece tokenizer. The model allows direct translation between any pairs of languages from among 100 languages. MBart50 supports up to 50 languages and similarly uses a shared vocabulary created with SentencePiece. However, it is trained on monolingual data in 50 languages, followed by multilingual fine-tuning with parallel data. Tower is a variant of LLama-2 that enhances the multilingual capability of the vanilla LLaMA-2 model for 10 selected languages. Its vocabulary follows LLaMA-2 without extension. 
In comparison to these models we test on, English-centric models like OpenAI's GPT exhibit significantly larger disparity in the tokenizer's vocabulary across different languages \cite{petrov_language_2023}, hence our attacks are likely to be better on them.

\smallskip
\noindent \textbf{Datasets.} 
For our dataset, we use the Flores dataset \cite{nllb2022}, which provides translations of the same sentence across various languages. This dataset facilitates the evaluation of translation quality using ground truth results available for numerous languages. Moreover, it supports testing across a range of different \{source language, target language\} pairs, ensuring that our results are generalizable.
The dataset is divided into training and testing sets, and we maintain the same split to ensure our results are reproducible. We profile output token density and output-to-input length ratio distributions from the training set with 1000 inputs per target language and attempt to infer the unknown target language in the testing set with 50 samples. Additionally, we use the ground truth translations to evaluate translation quality using BLEU scores for each target language.

\smallskip
\noindent \textbf{Languages.} 
For testing our attack, we selected target languages with the highest speaking populations, while ensuring sufficient diversity across different language families. The 13 languages we use for the target language to test our attack are shown in \cref{tab:language_families}.
For 
Note that Tower only supports 10 out of these 13 languages, and we exclude the unsupported languages (Hindi, Arabic, Japanese) for this model.
For M2N100 and Tower, we perform the testing using source languages of English, French, Spanish, and Russian. For MBart50, which needs fine-tuning for a particular source language, we only use English as the source language. 
We ensure all the models perform translation with sufficiently high accuracy, based on the BLEU score, as shown in Table~\ref{tab:translation_performance} in \cref{app:translation_validation}. 

\begin{table}[h!]
\centering
\caption{Target Languages Used to Test our Attack, Grouped by Language Family}
\begin{tabular}{|c|l|}
\hline
\textbf{Language Family} & \textbf{Languages} \\
\hline
Indo-European (Romance) & French, Spanish, Portuguese, Italian \\ \hline
Indo-European (Indo-Aryan) & Hindi \\ \hline
Indo-European (Germanic) & German, Dutch, English \\ \hline
Indo-European (Slavic) & Russian \\ \hline
Sino-Tibetan (Sinitic) & Chinese \\ \hline
Afro-Asiatic (Semitic) & Arabic \\ \hline
Japonic & Japanese \\ \hline
Koreanic & Korean \\ \hline
\end{tabular}
\label{tab:language_families}
\end{table}

\subsection{Results}

\subsubsection{Profiling}
\begin{figure}[h!]
  \centering
  \includegraphics[width=0.95\linewidth]{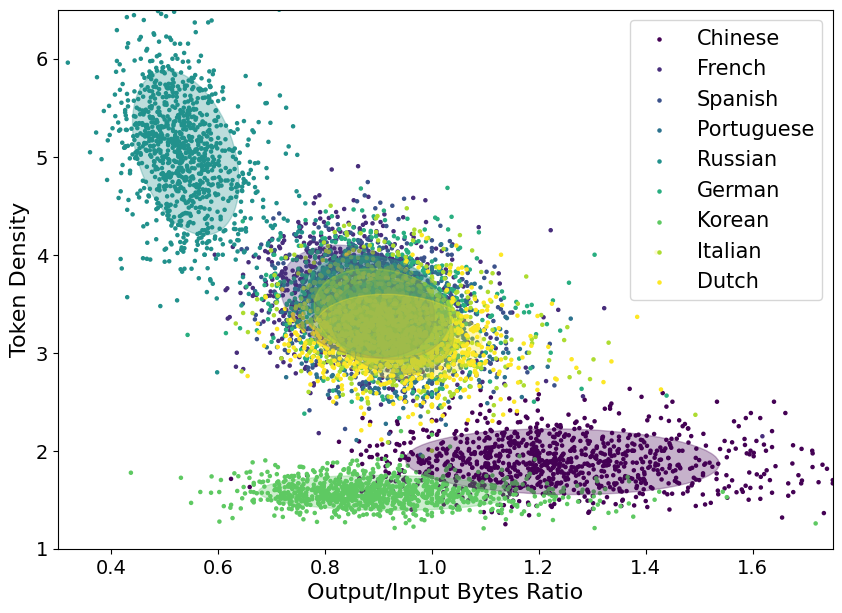}
  \caption{Profile using a Gaussian Mixture Model on a 2D decision space for the Tower model with the source language as English.}
  \label{fig:gmm_tower}
\end{figure}

\cref{fig:gmm_tower} shows a profile that we developed using the Tower~\cite{alves_tower} model, with the English source language and 10 different target languages, listed in \cref{tab:m2m100_source_languages} (excluding 3 languages that the model doesn't support). We profile 1000 inputs per language, repeating the same inputs for each language. The inputs are selected from the training split of the Flores-200~\cite{nllb2022} dataset. 
As shown in \cref{fig:gmm_tower}, both the Output Token Density and Output-Input Bytes Ratio play a distinct role in helping cluster datapoints of any given language. 
Particularly, resource-poor languages that have low token density (low bytes per token), such as Chinese and Korean, and those with distinctive morphology and Output/Input Bytes like Russian are easily identifiable in this profile.
Other languages can be identified based on the combination of the two metrics.
We construct similar profiles for each model across all source languages we test.

\subsubsection{Attack Success Rate}
We define the Attack Success Rate (ASR) of an attack on a translation task as the precision of the attacker's prediction about the target language, using 50 translation requests from the user. We average the ASR over 50 successive attacker predictions per language.

\begin{figure}[h!]
  \centering
  \includegraphics[width=\linewidth]{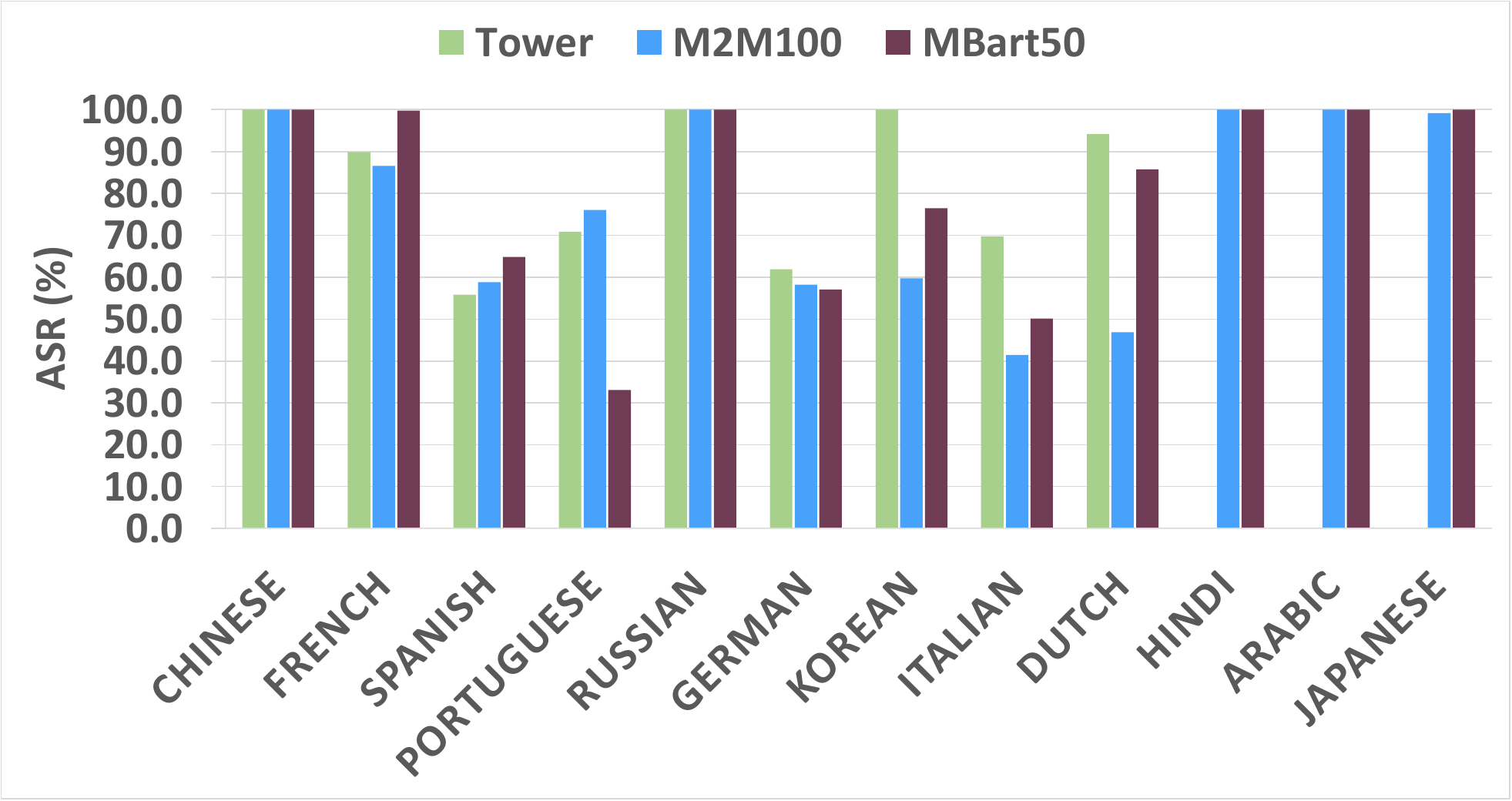}
  \caption{Attack success rates (ASR) for the attack leaking the output languages for different translation models, using source language as English.}
  \label{fig:translation_asr}
\end{figure}

\cref{fig:translation_asr} shows the ASR for our attack correctly guessing the target language on three different models, using English as the source language.
Overall our attack is highly effective, with the average ASR across all languages reaching 82.5\% for Tower, 77.3\% for M2M100 and 80.6\% for MBart50, respectively. 
Languages with more distinct morphology from different families, including Chinese, Russian and Korean, show 100\% precision for the Tower model. This is due to major differences in the byte encoding scheme, vocabulary sizes, and information density embedded in similar text lengths.
The lowest ASRs are achieved for languages coming from the Indo-European (Romance) family, where four languages are in the same family, where sometimes one language is misclassified as another in the same class.

\subsubsection{Additional Results}~\smallskip\\

\noindent \textit{Varying Attacker Samples.} In addition, we analyze the impact of varying the number of user requests the attacker samples to perform the prediction. While our default attack uses 50 requests from the user for the prediction, as the number of samples (user requests) used for prediction increases from 1 to 10 to 30 to 50, the ASR also increases on average from 48.0\% to 66.3\% to 76.5\% to 82.5\% respectively, as shown in \cref{fig:precision_tower_samples} in \cref{app:translation_extra_analysis}.

\smallskip
\noindent \textit{Alternative Source Languages.}
While we perform our attack using English as the default source language, our attack generalizes to other source languages as well. 
For Tower model, our ASR remains high across other source languages such as French (92\%), Spanish (93.3\%) and Russian (77.7\%).
Similarly for M2M100 model, our ASR is sufficiently high across other source languages such as French (81.3\%), Spanish (80.4\%), Russian (75\%). 
\cref{tab:tower_source_languages} and \cref{tab:m2m100_source_languages} provide the ASRs across all target languages for each of these source languages for Tower and M2M100 respectively in \cref{app:translation_src_languages}.

\smallskip
\noindent \textit{Alternative Test Dataset.}
We also demonstrate that the attacker does not need to specifically profile on the same data that the user uses for translation. 
For instance, we perform the same attack, where the user's translations (test dataset) are are selected from the  EuroParl~\cite{koehn-2005-europarl} translation dataset. The profiling still uses the Gaussian Mixture Model (GMM) trained with the Flores-200 training dataset. We show that even when the test set is the EuroParl dataset, the attack achieves comparable ASR (79.1\%) for the Tower model, as the ASR on the Flores-200 testing set (82.5\%). The ASR results across languages are provided in \cref{fig:precision_tower_europarl} in \cref{app:translation_extra_analysis}.

\subsubsection{Ablation Study}
\begin{figure}[h!]
  \centering
  \includegraphics[width=\linewidth]{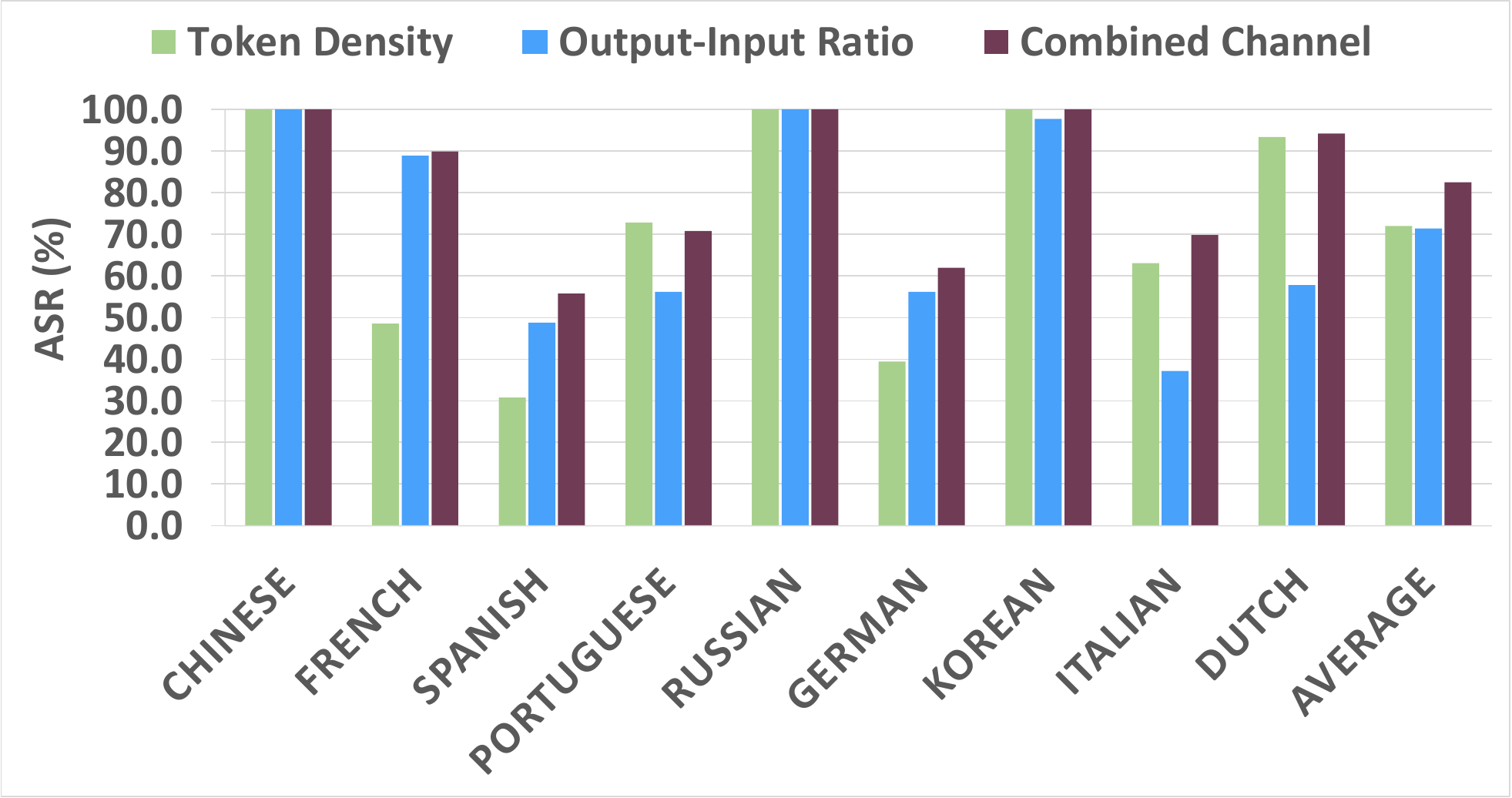}
  \caption{Ablation study showing ASR (\%) when only output token density or output-input bytes ratio is used as the side channel, compared with attack using combined channel, for Tower model with source language English.}
  \label{fig:translation_asr_ablation}
\end{figure}

In addition, we conducted an ablation study by including only one out of the two metrics for the attack, i.e., only the output token density or the output-input length ratio, and compared this against an attack using both these metrics (combined channel).
\cref{fig:translation_asr_ablation} compares these metrics for the Tower model. 
The results show that token density contains slightly more information than the output-input length ratio; however both attributes contribute towards increasing the overall precision in the combined channel. 
For instance, individually both token density and output-input length ratio alone only achieve an average ASR of around 70\%, whereas together they achieve an ASR of 82.5\%. 
Our ablation studies for the other models, such as M2M100 (\cref{fig:m2m100_ablation}) and MBart50 (\cref{fig:mbart50_ablation}), are shown in \cref{app:translation_ablation} and provide similar insights.

These results show that our attack can predict the language of LLM outputs even when a user is performing non-translation tasks, simply by analyzing the token density. Since the token-density channel is a fundamental property of the LLM and such attacks achieve high average precision (70\%), this potentially puts a user's language preference at risk of being leaked out across any task across most LLMs, highlighting the significance of our attack.

\section{Attack on Few-Shot Classification Tasks}
\label{sec:attack_on_prompt}
\subsection{Application Scenario}

\textbf{Context.}
LLMs are increasingly being used for text classification tasks, such as Amazon's Rufus~\cite{chilimbi_technology_2024} categorizing customer inquiries into product groups for personalized recommendations, Salesforce's Einstein~\cite{salesforce_einstein} organizing work summaries into issues for agent reviews, and Google MedPalm \cite{google_med-palm_nodate} generating medical diagnoses with explanatory reasoning. In these applications, explanations accompanying output classes improve performance \cite{ye2022unreliability}, enable human auditing, and reduce hallucinations. This work investigates whether the length of explanations in classification tasks correlates with the output class, potentially leaking class information through an output token count side-channel. We further explore the impact of biases in explanation lengths within few-shot prompting examples~\cite{brown2020fewshot}, a technique that uses input-output examples to guide response generation, to determine if biases in few-shot examples exacerbate output length variations and enhance side-channel leakage.

\textbf{Attack Setup.} We focus our attacks on an LLM performing binary text classification using the Predict-then-Explain (P-E) format \cite{ye2022unreliability}, where the output consists of predicted class followed by explanation; the output is one of two classes. We assume the attacker knows the specific task run on the LLM, but not the few-shot examples or the user's input to the LLM. 
The goal of the attacker is leak the output class for a given input to the LLM, as this often involves private information about the user, such as shopping behaviour~\cite{chilimbi_technology_2024}, work content~\cite{salesforce_einstein}, and medical condition~\cite{google_med-palm_nodate}.
As in the previous section, we assume the attacker has direct access to the exact output token counts. In \cref{sec:evaluation}, we show how this attack can be launched just by observing the execution time, remotely over the internet.

\subsection{Attack Design}
\textbf{Overview.} In classification tasks, LLMs may exhibit inherent biases to produce longer explanations for specific classes. Moreover, if the  explanations in the few-shot examples align with this bias, the differences in output token counts between classes can become even more pronounced. 
This variation creates a side-channel through which a malicious adversary can infer the classification result by observing or estimating the output token count. 
To leak information, the attacker must first characterize this behaviour by profiling the task's output token count distribution for a variety of inputs.
Since the output token count is also influenced by the input length, the attacker must also account for this.
Subsequently, the attacker can then observe the victim's execution, and by learning the output token count and input length and matching it to the profiled distribution, she can learn the victim's output class.
We perform the attack with both unbiased and biased few-shot examples, to evaluate the the natural potential for leakage in classification tasks via our token count side-channels and then the impact of few-shot prompting.

\textbf{Implementation.}
This attack has two-phases: a \textit{profiling phase} and an \textit{attack phase}, 
similar to previous attacks. 

\smallskip
\noindent \textbf{Profiling Phase.}
In the profile phase, the attacker uses the output token count, supplemented by the total input length in bytes, 
to establish a threshold for distinguishing between two predicted classes.
The attacker sends 200 inputs (100 per class) of known length to the LLM for classification, and observes the output class and the output token count for each response.
Then, the attacker fits an optimal threshold for the output token count, that can separate the two classes with the highest precision, defined as:
\begin{equation}\label{eq:thresh_prompt}
\textit{Threshold} = \alpha \times \textit{Input Length in Bytes} + \beta
\end{equation}
The optimal threshold is the one that achieves the optimal attack success rate (ASR) in the profiling phase under the constraint that the prediction precision for both classes is similar (e.g. 72\% vs 68\%, instead of 90\% vs 50\%).
We define the optimal ASR as follows, where $\theta$ is a hyperparameter to choose how close the precision between classes should be (in the experiment 0.5):
\begin{equation}
\textit{Optimal ASR} = \frac{(\textit{Prec.1} + \textit{Prec.2})}{2} - \theta|\textit{Prec.1} - \textit{Prec.2}|
\label{eq:optimal_asr}
\end{equation}

\smallskip
\noindent \textbf{Attack Phase.}
The attacker monitors the (1) input length in bytes and the (2) output token count of a user's classification request, and determines whether the result falls above or below the threshold, thereby inferring the predicted class.

\subsection{Experiment Setup}

\textbf{Datasets.} To get representative classification tasks, we use the Natural Instructions Dataset \cite{supernaturalinstructions}, which consists of 61 distinct tasks, their human-authored instructions and 193,000 inputs. 
From this data set, we select all 12 binary classification tasks for our experiments, as shown in \cref{tab:fewshottasks} in the Appendix. While the dataset includes tasks and inputs and outputs, it does not provide explanations for the output, which we need for our few-shot examples. For this reason, we create a synthetic dataset of explanations using these inputs and outputs with state-of-the-art, GPT-4-turbo \cite{openai2024gpt4technicalreport}, which we then use as few-shot examples. 

\begin{figure}[h!]
  \centering
  \includegraphics[width=0.95\linewidth]{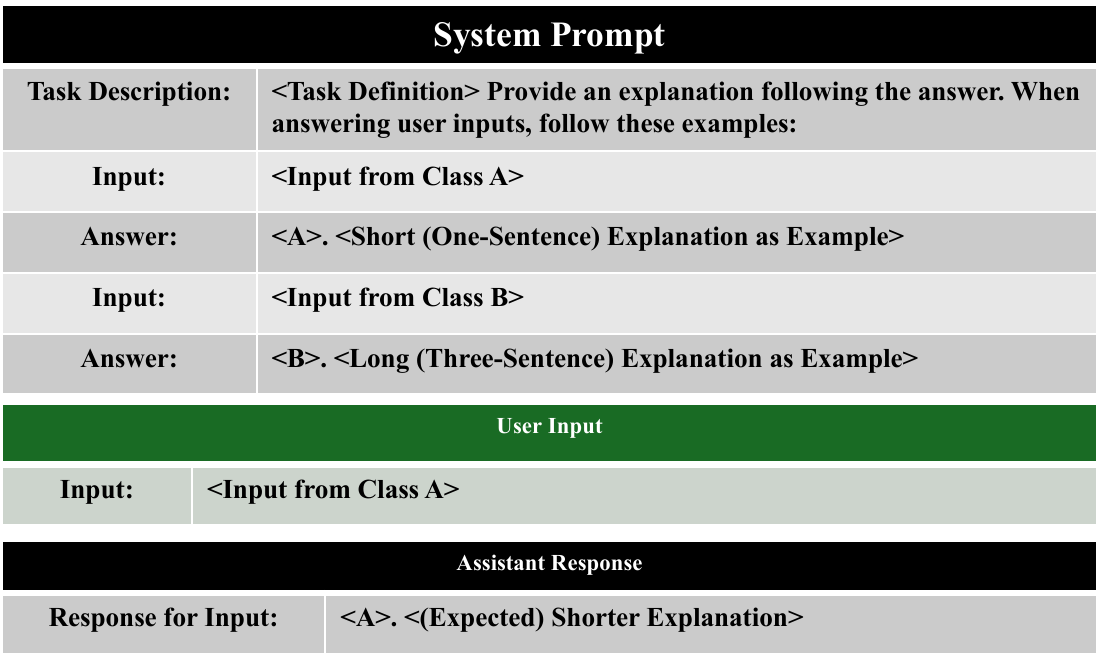}
  \caption{Prompt template for our tasks, using the Predict-then-Explain (P-E) format. The system prompt consists of the task description and the input examples for each class (e.g., one-shot example per class). If the few-shot examples have biased explanation lengths (shorter for class A), it is possible the response with output class A also has a shorter explanation.}
  \label{fig:one-shot-biased}
\end{figure}

\textbf{Prompt.} The system prompt to the LLM consists of a task description from the Natural Instruction dataset, and an instruction requesting an explanation, and a set of few-shot examples that include predicted labels and explanations in the P-E format. The user prompt then provides the task input, drawn from the ``inputs'' field of the natural-instruction dataset. The final output generated by the language model will follow the same P-E format as the few-shot examples.
As prior work suggests using a balance of examples across different classes for the best performance\cite{zhao2021calibrate}, we use an equal number of few-shot examples per class: one, two, or three few-shot examples per class, which we refer to as one-shot, two-shot, and three-shot respectively.
\cref{fig:one-shot-biased} shows the template for our \textit{one-shot} prompt to the LLM.

\textbf{Experiments.} To characterize the biases in the explanation lengths between classes of a given task, we provide each input class either a \textit{short}, one sentence explanation, or a \textit{long}, three sentence explanation in the few-shot examples.
Thus, we have four categories of examples for the two classes, long-long, and short-short which are \textit{unbiased}, and short-long and long-short that are \textit{biased}. 
Using the unbiased examples, whic have explanations of equal length  for both classes, we characterize the inherent bias in explanation lengths generated for a given task.
For biased examples, we categorize them on a per task level as \textit{augmenting} or \textit{diminishing}, based on whether their bias is the same or opposite direction as the inherent bias in the task. 
With these examples, we can characterize the diminishing or augmenting effect of biased examples on the output explanation lengths and observe the best/worst-case leakage of output class from token counts for a given task. 
We use 200 inputs (100 per class) for training profiles and  200 inputs (100 per class) for testing our attack per category.

\smallskip
\textbf{Models.} We evaluate open-source models from the Gemma-2 \cite{gemmateam2024gemma2improvingopen} and Llama-3 \cite{llama3modelcard} family, as well as proprietary models from the GPT-4o \cite{openai_2024_gpt4o} family. We select these due to their state-of-the-art accuracy in classification tasks.

\subsection{Results}

\subsubsection{Profiling Explanation Length Bias}

\begin{figure}[h!]
  \centering
  \includegraphics[width=0.9\linewidth]{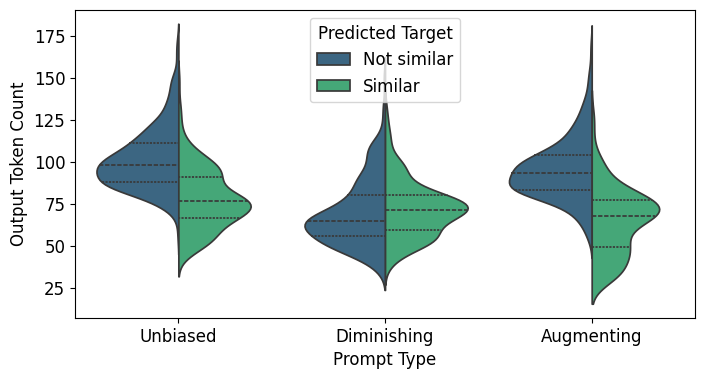}
  \caption{Output Token Distribution for Task-145, with Unbiased, Augmenting and Diminishing examples used as few-shot examples (three-shot per class). The Unbiased is chosen as ``Long-Long''.}
  \vspace{-0.1in}
  \label{fig:violin}
\end{figure}

To characterize the inherent bias in explanation lengths in tasks, and also the best/worst-case biases due to biased few-shot examples, 
we characterize each of the tasks with unbiased (long-long), and biased examples of the augmenting and diminishing types and measure the output token counts for the two output classes of the task. 
\cref{fig:violin} shows the visualization of the output token counts with the Gemma2-9B model for one such task (Task-145) that classifies a pair of arguments into `Similar' or `Not Similar', based on whether they support or attack the other's stance, with three-shot per class.

For the unbiased examples, the inherent bias of Gemma2 causes it to generate longer responses for the `Not Similar' class, and shorter ones for `Similar'. With \textit{augmenting} examples (long-short), the difference is amplified creating a considerable separation in token-counts for the two classes. On the contrary, \textit{diminishing} examples (short-long) cause the bias to move in the opposite direction with the classes having a similar distribution of token counts, and in fact the `Not-Similar` class now has a shorter explanation.
This validates the fact that LLMs have natural explanation length biases based on output classes, and these can get augmented or diminished (or biased in the opposite direction) based on the biases in the example explanation lengths.

Using \cref{eq:thresh_prompt} and \cref{eq:optimal_asr}, we derive the thresholds for each of the tasks with each model and few-shot example category, which enables our attack.

\subsubsection{Attack Success Rate}

\begin{figure}[h!]
  \centering
  \includegraphics[width=\linewidth]{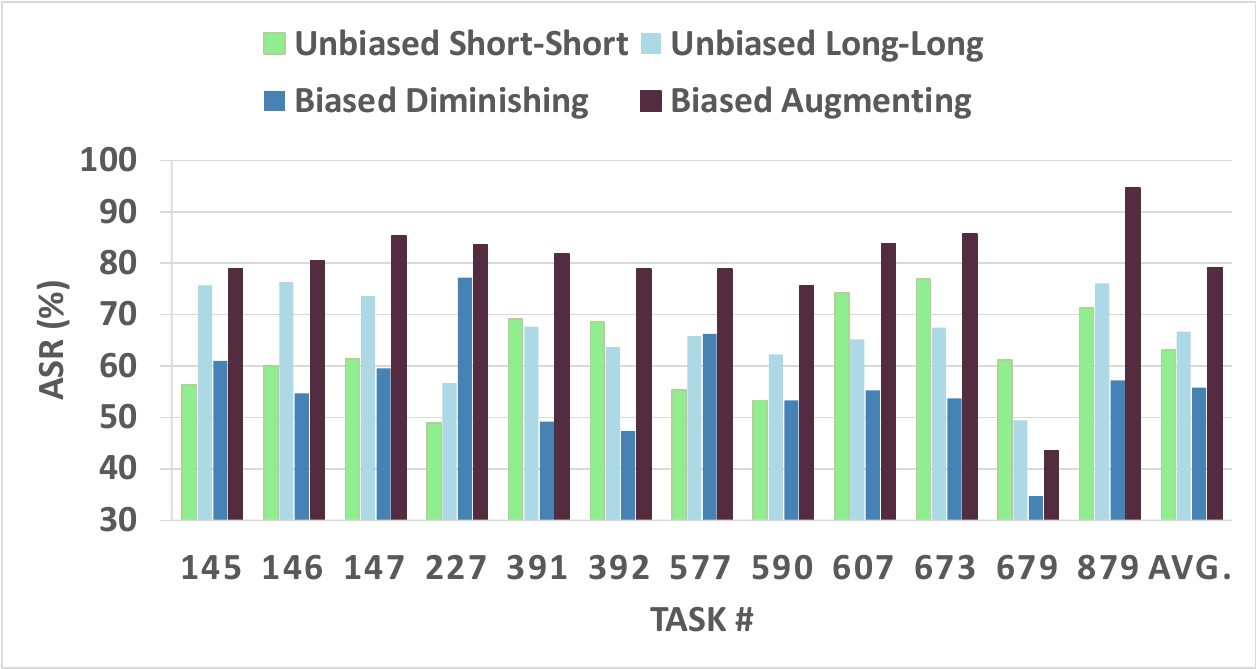}
  \caption{ASR (\%) across all tasks for Gemma2-9B, with unbiased and biased examples with 3-shot examples per class.}
  \label{fig:gemma_biased}
\end{figure}

\cref{fig:gemma_biased} shows the ASRs (measured as the average precision for predicting each class) across all 12 tasks for the Gemma2-9B model with 3-shot examples per class. With the baseline unbiased examples, we see that the average ASR reaches 63.1\% (short-short) and 66.7\% (long-long), with some tasks having ASR higher than 75\%. This is 
much higher than the ASR with random guesses (50\%), indicating that the output token counts are naturally biased in tasks even without any biased examples. When the task is prompted with \textit{augmenting} few-shot examples, the distribution shifts even more towards the inherent bias, leading to an average 15\% increase in the attack success rate. In the \textit{diminishing} direction, the results are mixed. Either the explanation exhibits a less significant length difference in the opposite direction, or the explanation bias is still in the same direction as the inherent bias to a smaller degree. The diminishing examples result in a best-case ASR of 75\% (Task 227), and an average ASR of 55\%.

Task-679 is an outlier where all of the categories (unbiased or biased) have low ASR (around 50\%). This is because the output token count for this task does not have a strong bias for either of the classes, making our attack ineffective. With biased examples, it tends to shift a subset of explanations to a disproportionally long length, making the current threshold less stable.
However, for most tasks, overall, our side-channel attack is quite successful in leaking output class based on output token counts. 

\begin{table*}[h!]
\centering
\caption{ASR (\%) of Augmenting class prompts for different model architectures and sizes}
\begin{tabular}{|c|c|c|c|c|c|c|c|}
\hline
\textbf{} & \textbf{Gemma2-2B} & \textbf{Gemma2-9B} & \textbf{Gemma2-27B} & \textbf{Llama3.1-8B} & \textbf{Llama3.2-3B} & \textbf{GPT-4o} & \textbf{GPT-4o-mini} \\ \hline
\textbf{391} & 79.2 & 81.8 & 78.3 & 83.3 & 70.5 & 80.3 & 92.5 \\ \hline
\textbf{227} & 70.5 & 83.5 & 72.2 & 83.8 & 73.8 & 91.0 & 94.2 \\ \hline
\textbf{392} & 67.8 & 78.8 & 83.8 & 86.0 & 69.7 & 82.3 & 94.2 \\ \hline
\textbf{590} & 70.3 & 75.7 & 63.5 & 69.7 & 66.2 & 87.8 & 87.0 \\ \hline
\textbf{879} & 89.3 & 94.7 & 82.1 & 93.5 & 81.5 & 93.3 & 97.7 \\ \hline
\textbf{Average} & \textbf{75.4} & \textbf{82.9} & \textbf{76.0} & \textbf{83.3} & \textbf{72.3} & \textbf{86.9} & \textbf{93.1} \\ \hline
\end{tabular}
\label{tab:prompting_model_sizes}
\end{table*}

\subsubsection{Varying Number of Few Shot Examples} 
\begin{figure}[h!]
  \centering
  \includegraphics[width=\linewidth]{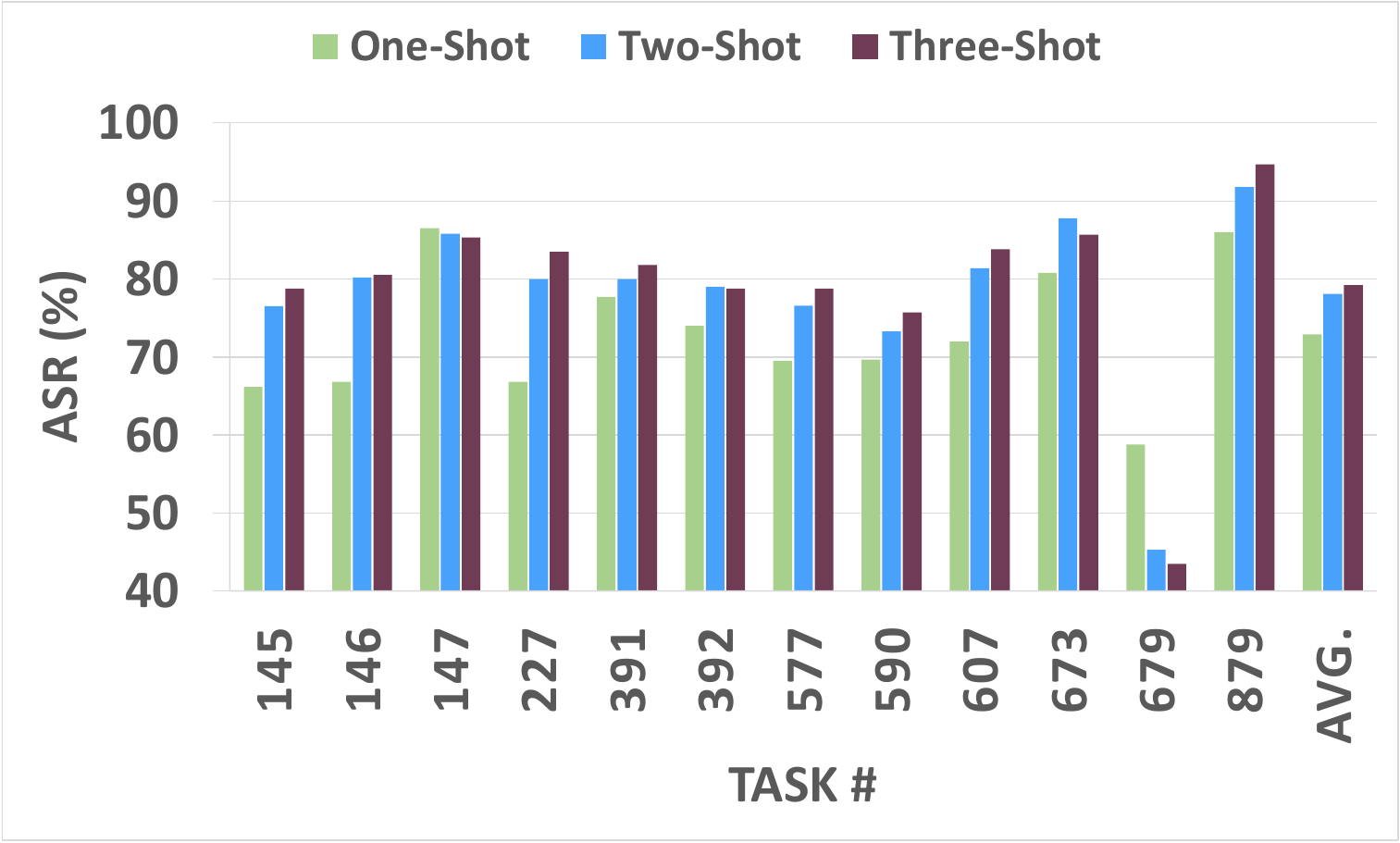}
  \caption{ASR (\%) as the number of few-shot examples is varied from 1-shot to 2-shot to 3-shot, for Gemma2-9B, with augmenting examples.}
  \label{fig:gemma_demo}
\end{figure}

\cref{fig:gemma_demo} shows the ASR as the number of few-shot examples per class is varied, for the \textit{augmenting} examples, for all the tasks with Gemma2-9B. On average, the ASR increases significantly from one-shot (72.9\%) to two-shot (78.1\%), and then shows a moderate increase to three-shot (79.2\%). This shows that the biases in output token counts for most tasks gets accentuated, the more biased examples are added, although this effect has diminishing returns.

\subsubsection{Alternative Model Architectures and Sizes}
To validate that the biases persist across different LLMs and whether there is a correlation with the model size, we test our attacks on several state-of-the-art LLMs, Gemma-2, LLaMa-3 and GPT-4o models and different model sizes. Due to the large run times for these experiments, we only tested a subset of 5 representative tasks across all models with the \textit{augmenting} examples and 3-shot prompts. \cref{tab:prompting_model_sizes} shows the results.

We observe that GPT-4o family has the highest ASR, with GPT-4o with average ASR of 86.9\% and GPT-4o mini (93.1\%); this is due to strong bias in the output token counts. In comparison, LLaMa-3.1 and 3.2 have average ASR of 83.3\% and 72.3\% respectively, and Gemma-2 models have average ASR between 75.4\% to 82.9\%. There is no strong correlation with model size. While GPT-4o-mini shows more bias than its larger counterpart GPT-4o, Llama-3.1/3.2 shows opposite behaviour with less bias (lower ASR) for the smaller model size, and Gemma2 has similar lower ASR for the smallest (2B) and the largest model (27B) sizes.

\subsubsection{Ablation studies}~\\

\noindent \textit{Model Temperature.}
A higher model temperature introduces randomness in generated outputs and may add noise in the output token distribution that impacts out attack. Our default experiments were done with a temperature of 0 (or 0.0001 when a zero temperature is not supported by the model) for reproducibility. To test the impact of temperature, we vary the temperature from 0 to 0.3 to 0.7 and evaluate the ASR (with augmenting 3-shot examples) for the GPT-4o and Gemma-2 models.
For GPT-4o, the ASR drops from 82.9\% to 82.1\%, to 80.8\%, whereas for Gemma-2, it varies from 86.9\% to 87.1\% to 87.8\%, as the temperature goes from 0 to 0.3 to 0.7. 
Thus, our attack is highly resilient even under relatively high temperatures for classification tasks. 
\cref{tab:asr_fewshot_temperature} in the \cref{app:abl_classification_attack} provides the ASR results across all tasks.

\smallskip
\noindent \textit{Correct Model Predictions.}
Although model accuracy is not our primary focus, we wish to examine whether correct model class predictions  have correlation with the length of generated output explanation. To study this, we filter the results to only include cases where the predicted target matches the ground truth. With this filtering, the average ASR increases slightly by 1\%, however, this change is not substantial to suggest any correlation between correct predictions and bias in output lengths. \cref{tab:asr_correct_incorrect} in \cref{app:abl_classification_attack} shows task-wise ASRs for Gemma2-9B and GPT-4o.

\section{End-to-End Timing Side-Channel Attacks}
\label{sec:evaluation}
\subsection{Network Scenario}
In this section, the attacker no longer has perfect access to the generated output token counts. Instead, the attacker is located on the network between the application server and the victim, and has the ability to monitor the network packets to obtain the length of the input text, and the timing between the input and output packet. The LLM is assumed to operate in a non-streaming mode, where it sends the entire output back after it is completely generated, and the batch size is set to one such that it will only process one request at a time. We also assume that the attacker has access to the model as a normal user so that she can profile the timing of inputs with different classes, such as the target language and classification results. Additionally, the attacker is assumed to have a network connection similar to that of the victim.

\subsection{Attack Method}
The attack takes place in two phases similar to previous attacks: a profiling phase and an attack phase. In the profiling phase, the attacker sends the requests with knowledge of the sensitive attribute (i.e. either language or output class) inside the input and sends inputs repeatedly to obtain measurements for each input class. This process, called ``class profiling'', extracts the characteristics of each input class. It mirrors the profiling phase discussed in earlier sections, except this time, the token count is inferred from timing information. Additionally, the attacker performs ``network profiling'' to account for the effect of time-dependent latencies due to network congestion and server-side workload interference. We study the frequency at which an attacker has to perform network profiling, especially for commercial LLMs, who's latencies may vary rapidly due to the workloads of other customers. In this case, the attacker may also perform ``concurrent network profiling'', which involves sending requests every minute to measure TTFT (Time to First Token) and TPOT (Total Processing Time) to estimate system and network delays, as well as the time required to generate each output token. To mitigate outliers, the our attacker uses a 5-minute window to and takes the median of five TTFT and TPOT measurements, thus smoothing out large variations due to changes in server load.

In the attack phase, the attacker will aforementioned network profiling results to adjust the timing information before using the class profiles to predict the sensitive attribute.

\subsection{Token Recovery Error}
\label{sec:token_recovery_error}
We first confirm the linear relationship between the number of generated tokens and the generation time of the LLMs when running on a local machine. We conducted a series of experiments on a machine equipped with an Nvidia RTX A6000 GPU, AMD Ryzen Threadripper PRO 5945WX, and 32GB of RAM for all the open-source models used in our previous experiments. For the proprietary LLM GPT-4o, timing measurements were taken from the moment a request was sent via the OpenAI API until a response was received. We use the Pearson correlation coefficient to measure how well generation time linearly correlates with the number of generated tokens. For models running on a local machine, the Pearson coefficient is at least 0.987, with the Tower model achieving a nearly perfect linear correlation (See \cref{tab:token_error_rate}). These results demonstrate a strong linear relationship and confirm that the input processing time is negligible.  In contrast, GPT-4o exhibited a significantly weaker correlation, with a coefficient of only 0.370. The lower correlation of GPT-4o could be due to a number of factors which we speculate to include optimization techniques, such as speculative decoding or throughput enhancements, which could disrupt the relationship between token count and generation time. In addition, as mentioned above, unpredictable network congestion and interference from other requests may lead to delays in processing our GPT-4o requests. As we show below, many of these unknown source of profiling noise can be mitigated with concurrent profiling.

\begin{table}[h!]
\caption{Perason Correlation Coefficient between Token Count and Generation Time}
\centering
\begin{tabular}{|c|c|c|c|c|c|}
\hline
& \textbf{Tower} & \textbf{M2M100} & \textbf{MBart50} & \textbf{Gemma2} & \textbf{GPT-4o} \\ \hline
\textbf{Pearson} & 1.000 & 0.990 & 0.989 & 0.987 & 0.370 \\ \hline
\end{tabular}
\label{tab:token_error_rate}
\end{table}

\subsection{Translation Workload}
As the intrinsic nature of autoregression suggests, an approximate linear relationship exists between the number of generated tokens and the execution time. We have:
\begin{equation}
\textit{\# of Generated Tokens} \propto t
\end{equation}
Since we don't need the exact number for token density but rather their relative relationship, we can effectively recover this relationship by approximating it with the execution time. Therefore, the token density calculation can be expressed as:
\begin{equation}
\textit{Token Density} = \frac{L_{\text{output}}}{\textit{\# of Generated Tokens}} \propto \frac{L_{\text{output}}}{t}
\end{equation}
where $L_{\text{output}}$ represents the length of the output.
For the attack on translation workload, we chose the models as used in our previous evaluation, including M2M100, Tower, and MBart50. We implement the client side on an AWS t3.large instance located in Ohio, which sends translation requests of sentences to the target language. The server side, which hosts LLM on a g6.xlarge instance with NVIDIA L4 Tensor Core GPUs, is located in Oregon. The client first sends the target language and the sentence to be translated to the server. After performing the translation, the server responds with the translated text in plaintext. We estimate the round trip time using a ping request and subtract this time from the measured latency to estimate the execution time of the LLM. We 
compare the precision of recovering the target language using timing versus having access to the actual output token count in \cref{tab:translation_network}. Note a positive value in the brackets means that recovery accuracy with timing was actually \textit{higher} than with the actual output token count. We attribute slight differences (i.e. less than 5\%) to measurement error---the only one with noticeable performance change is with German on the M2M100 model where recovery with timing experienced a 14.2\% ASR drop. %

\begin{table}[h!]
\centering
\caption{Language prediction precision in the network scenario with timing information for 50 samples (Difference between timing and output token precision shown in brackets)}
\begin{tabular}{|c|c|c|c|}
\hline
\textbf{Language} & \textbf{Tower} & \textbf{M2M100} & \textbf{MBart50} \\ \hline
Chinese     & 100.0(+0.0) & 100.0(+0.0) & 100.0(+0.0) \\\hline
French      & 90.7(+0.8)  & 83.4(-3.1)  & 100.0(+0.2) \\\hline
Spanish     & 51.8(-4.0)  & 63.5(+4.6)  & 67.8(+2.9) \\\hline
Portuguese  & 72.8(+1.9)  & 75.3(-0.8)  & 34.0(+1.0) \\\hline
Russian     & 100.0(+0.0) & 100.0(+0.0) & 100.0(+0.0) \\\hline
German      & 63.1(+1.1)  & 44.0(-14.2) & 58.8(+1.8) \\\hline
Korean      & 100.0(+0.0) & 60.4(+0.7)  & 77.6(+1.2) \\\hline
Italian     & 73.4(+3.6)  & 38.7(-2.7)  & 50.3(+0.2) \\\hline
Dutch       & 97.0(+2.9)  & 45.4(-1.5)  & 87.7(+2.0) \\\hline
Hindi       & -           & 100.0(+0.0) & 100.0(+0.0) \\\hline
Arabic      & -           & 100.0(+0.0) & 100.0(+0.0) \\\hline
Japanese    & -           & 98.5(-0.7)  & 99.7(-0.3) \\\hline
\textbf{Average} & \textbf{83.2(+0.7)} & \textbf{75.8(-1.5)} & \textbf{81.3(+0.7)} \\\hline
\end{tabular}
\label{tab:translation_network}
\end{table}

\subsection{Classification on Open-source Models}
We first conducted our evaluation using open-source models from the Gemma2 family. We selected the Gemma2-9b-IT model, which was executed on a local machine with the same spec as in \cref{sec:token_recovery_error}. The results are presented in \cref{tab:gemma9b_x_demo_time}. Our analysis shows that the ASR drop in all tasks is relatively minor, with an average variation of only 3.5\% in ASR. This finding indicates that the timing signal closely resembles the output token count, making the timing side channel a viable means of leaking sensitive attributes. The most significant drop for ASR came with task607, which had significant class imbalance and resulted in \cref{eq:optimal_asr} selecting a threshold that balanced for precision in both classes instead of an optimal ASR. 

\begin{table}[h!]
\centering
\caption{ASR (\%) of Gemma-9B-IT with Augmenting class prompts with timing information}
\begin{tabular}{|c|c|c|c|}
\hline
\textbf{Task \#} & \textbf{One-Shot} & \textbf{Two-Shot} & \textbf{Three-Shot} \\ \hline
145     & 57.8(-8.4)  & 64.2(-12.3) & 69.0(-9.8) \\\hline
146     & 63.5(-3.3)  & 72.2(-8.0)  & 72.3(-8.2) \\\hline
147     & 86.3(-0.2)  & 85.2(-0.6)  & 85.5(+0.2) \\\hline
227     & 73.0(+6.2)  & 79.5(-0.5)  & 79.0(-4.5) \\\hline
391     & 77.0(-0.7)  & 78.8(-1.2)  & 81.0(-0.8) \\\hline
392     & 74.0(+0.0)  & 78.7(-0.3)  & 79.0(+0.2) \\\hline
577     & 69.4(-0.1)  & 77.9(+1.3)  & 79.7(+0.9) \\\hline
590     & 71.2(+1.5)  & 73.2(-0.1)  & 78.3(+2.6) \\\hline
607     & 65.3(-6.7)  & 64.7(-16.7) & 62.1(-21.7) \\\hline
673     & 80.8(+0.0)  & 87.3(-0.5)  & 85.7(+0.0) \\\hline
679     & 58.6(-0.2)  & 44.1(-1.2)  & 43.5(+0.0) \\\hline
879     & 86.3(+0.3)  & 92.5(+0.7)  & 93.9(-0.8) \\\hline
\textbf{Average} & \textbf{71.9(-1.0)}  & \textbf{74.9(-3.3)}  & \textbf{75.8(-3.5)} \\\hline
\end{tabular}
\label{tab:gemma9b_x_demo_time}
\end{table}

\subsection{Classification on GPT4-o}

As discussed in \cref{sec:token_recovery_error}, the proprietary GPT4-o exhibits significantly more variation in delays than our open-source models. We thus study the effectiveness of three different profiling methods. The first method involves measuring the time between sending a request to the OpenAI API and receiving a response. This delay is then assumed to be linearly correlated with the output token count and doesn't consider other sources of timing variation. We consider this na\"ive method our baseline for these experiments. The second method uses uses profiles obtained over a period of time and uses their average to estimate the output token count, thus assuming a that timing variations are small and can be approximated by a single average factor. The third and final method utilizes concurrent profiling by sending multiple requests to monitor the Time to First Token (TTFT). This measurement helps us estimate network and system delays concurrently with the attack, as well as the Time per Output Token (TPOT). Measuring network and system delay concurrently enables the attacker to account for varying server loads throughout the day. By combining these two pieces of information, we can estimate the number of output tokens for a request made at a similar time during the concurrent profiling session.

Class profiling is less sensitive to timing variation and need not  be performed on the same day as the attack. For the class profile to be used on all tests, we took profiles on all days and chose the one with the best discriminator (i.e. highest ASR on the profile data), which happened to be the profile taken on Day 4.

The GPT-4o model, as we measured, has higher errors in recovering output token count using timing due to various server loads and network conditions. We use a concurrent network profiler that is located in the same city but in different locations and networks as the user. As of our measurement, the network latency remains relatively stable, which only varies by roughly $\sim$30\% when excluding outliers (see \cref{fig:network_delay}). While the time per output token can vary more than 2 times even when extreme outliers are excluded (\cref{fig:time_per_token}). This suggests that timing variability is caused more by server-side variations rather than network latency.

\begin{figure}[h!]
  \centering
  \includegraphics[width=\linewidth]{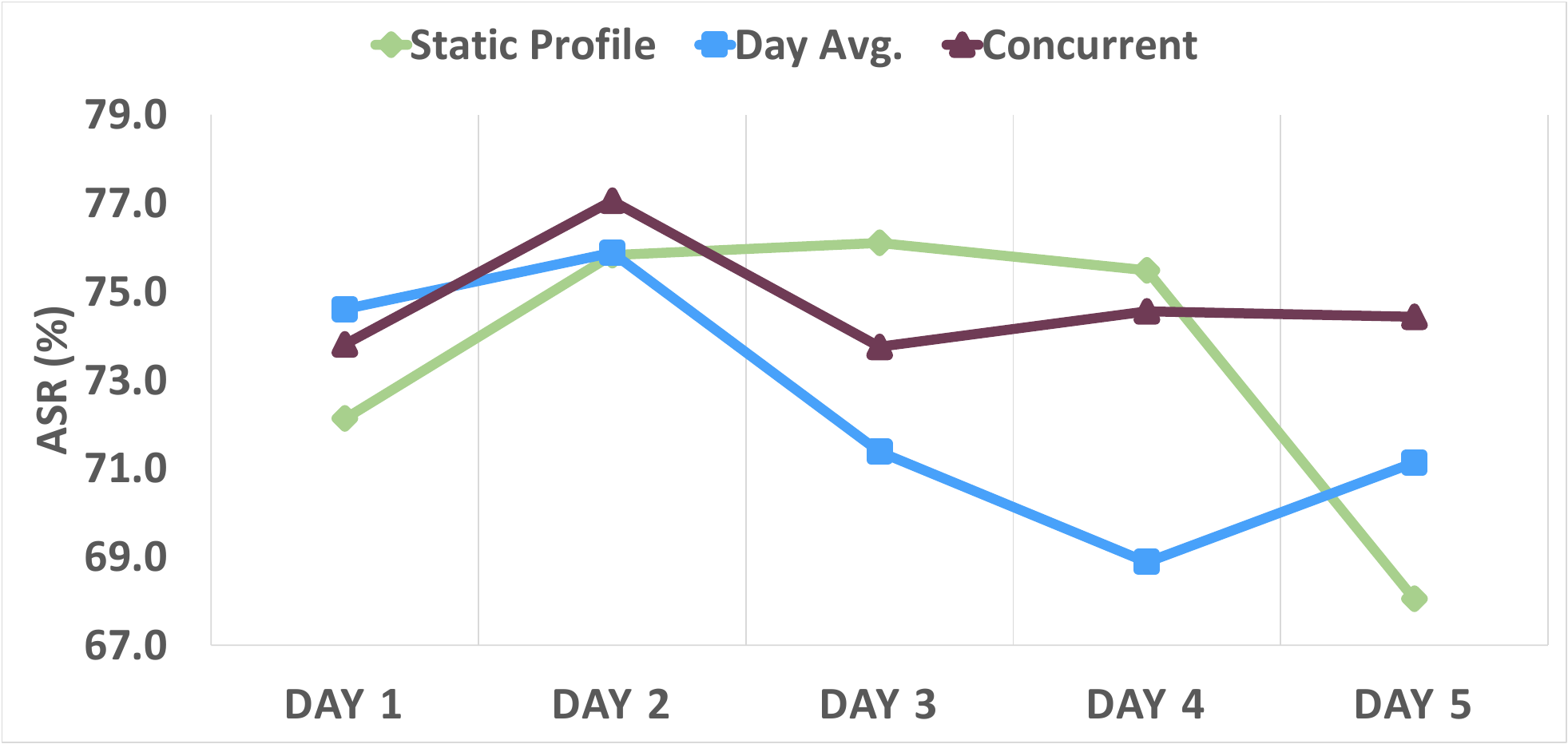}
  \caption{Average ASR (\%) across different tasks for remote timing attack on GPT-4o on different days}
  \label{fig:network_chart_line}
\end{figure}

The average ASR across 5 days for different network profiling technique is shown in \cref{fig:network_chart_line}. The concurrent profiling technique shows the highest ASR on average and behaves more stably across different days. Most notably, concurrent network profiling achieves an ASR of 6\% higher on Day 5, representing better tolerance for different network conditions. When the network and server load remains relatively stable, concurrent has a similar performance to the other two methods. Using concurrent profiling, the average ASR on GPT4-o is 0.9\% lower than the average ASR on our open-source models.

\section{Potential Defences}
\label{sec:defence}
\subsection{Tokenizer-level Defence}
To address the issue of token density differences among languages, it is crucial to focus on the root cause, which lies in tokenization. Previous efforts to mitigate language bias, such as allocating language capacity based on average log probabilities to optimize downstream performance \cite{zheng2021allocating}, primarily target performance improvements but do not guarantee consistent token density across languages. To effectively mitigate the timing channel, achieving a more uniform token density is key, thus making the target language anonymous. Despite recent efforts to create more balanced training datasets, intrinsic differences between languages often require varying vocabulary sizes in tokenizers. While changes in the tokenization process may mitigate this side channel, such modifications raise important questions about their potential impact on model performance. We leave this line of investigation for future work.

\subsection{Prompt-level Defence}
For the classification scenario, when providing explanations, one possible solution is to instruct the model to generate a fixed-length output through the prompt. We tested whether this approach could effectively reduce side-channel leakage without altering the demonstration examples by modifying the system prompt from ``Provide an explanation'' to ``Provide a 60-word explanation,'' where 60 was selected as the median word count for three-sentence explanations. To evaluate this defense, we pick highly vulnerable scenarios with the high ASR: three very biased tasks, task227, task391 and task879, using both Gemma2-9b-IT and GPT-4o models prompted with augmenting three-shot examples. We assume the attacker has perfect access to the output token count, giving an average  ASR with no defense of 79.3\% and 88.2\%, respectively.

Keeping the three-shot examples unchanged, we tested whether modifying the prompt alone was sufficient to mitigate the side channel. The two models displayed distinct results: the originally more biased GPT-4o experienced a 26\% reduction in ASR, dropping to 61.9\%, while the Gemma2-9b-IT model maintained a similar level of bias, with a slight increase in ASR to 80.3\%. This experiment shows that the prompt-level defense is only effective on some models while completely ineffective on others. A viable option for models that do not respond to this prompt is to use diminishing few-shot examples to counter the natural bias of the LLM.

\subsection{System-level Defence}
For tasks such as machine translation, there is no straightforward way to alter the output token count without changing the tokenization methods. However, modifying the tokenization method for existing models typically requires retraining from scratch, which is often impractical. Therefore, one strategy is to pad the output token count and the output plaintext length length to match the longest expected output length. With these two metrics obscured, the output-input ratio and output token density will be maintained at a more or less constant value, effectively mitigating the side channel proposed in Section~\ref{sec:attack_on_translation}.

When delaying the faster responses, the highest latency penalty for this approach is observed in Tower, where the language with the highest average token count is Korean, which requires approximately twice as many tokens as other languages due to its unchanged tokenization vocabulary inherited from Llama-2. The average performance cost to mitigate the output token count side channel across languages is 93\%. In contrast, models like MBart50 and M2M100 exhibit closer average token counts across languages, resulting in an average performance penalty of around 13\%.

To achieve a similar level of output byte length, Tower would need to align with the plaintext length of Russian, necessitating an average of 67\% padding. MBart50 and M2M100 require an average of 108\% padding due to their additional support for Hindi, which has the longest average plaintext length.

\section{Discussion}
\label{sec:discussion}
\textbf{Impact of dynamic few-shot examples: }In the few-shot classification scenario, we use statically chosen examples that do not change based on the user's input. However, there is a growing trend to use Retrieval Augmented Generation (RAG) to improve model performance by dynamically selecting in-context examples. Under this scenario, we anticipate that if the underlying dataset is biased toward a particular class concerning explanation length, it will produce behaviour similar to that of the biased class prompt. Conversely, an unbiased dataset should perform similarly to the unbiased class prompt. An interesting scenario may arise when the dataset contains outliers that have exceedingly long or short explanations. The presence of such outliers may lead to yet another side-channel. We leave the exploration of this phenomenon as future work.

\textbf{Impact of latency and throughput optimizations: }For our evaluation, we assume a batch size of 1 to optimize latency and replicate the scenario of a low server load. In real-world systems, applications often aim to achieve low latency and high throughput. One notable batching technique is introduced in Orca \cite{280922}, which has been further refined into continuous batching that operates on an iteration level. Despite potential noise, we expect the observation channel to remain strong, as responses are still returned at an iteration level that maintains a linear correlation with the output token count. Other optimization techniques, such as speculative decoding, may disrupt the linear relationship to some extent. However, a rough linear relationship is likely to persist. A promising direction for future work would be to explore how latency optimization techniques might impact this observation channel.

\textbf{Other vulnerable applications:} In addition to the two scenarios evaluated in this paper, there may be other potentially vulnerable applications. For example, in a system that uses a prompt to enforce rule-based analysis of user input in natural language, when an LLM provides explanations for the rules that have been violated, differences in the number of violations could lead to varying response lengths. This variation might allow users to infer how many rules have been broken, thus leaking information about the rules embedded in the system prompt. The key criterion for identifying output token count side-channel vulnerabilities is whether changes to a specific attribute result in a measurable shift in the distribution of output token counts. Therefore, for LLMs designed for specialized tasks, developers should evaluate whether sensitive attributes can change the output token counts, as this has a risk of information leakage.

\section{Related Work}
\label{sec:related}

\subsection{Side-channel Attacks on LLMs}

\subsubsection{Side-Channels in LLM Pipelines}
Side-channels in LLMs were pioneered by Debenedetti et al. \cite{299864}, who identified vulnerabilities across various stages of LLM systems by exploiting multiple side-channel signals. Specifically, they uncovered weaknesses arising from training data filtering, input preprocessing, model output filtering, and query filtering. Of particular relevance to our work is their input preprocessing attack, which leveraged the tokenization process due to limitations in the context window size. Their approach leaked out uncommon strings present in the tokenizer's vocabulary. In contrast, our approach shifts focus to a higher-level analysis of linguistic morphology differences, and exploits biases in vocabulary resource allocations across languages to leak the language, rather than targeting leakage of specific strings in the tokenizer's vocabulary.

\subsubsection{Token Length Side-Channel} Recent LLM keylogging attacks \cite{299888} analyzed network packet sequences to infer the length of individual output tokens for LLMs operating in streaming mode. By determining token lengths and recognizing the tendency of LLMs to repeat training data, the authors fine-tuned an inference model to deduce the exact plaintext of the output response. In contrast, our attack works on even non-streaming mode LLMs, where the attacker cannot infer output token lengths. Our approach instead focuses on leaking sensitive input attributes based on the number of output tokens and the execution time and does not rely on individual token sizes.

\subsubsection{Side-Channels Due to Performance Optimizations} 
Wei et al.~\cite{wei2024privacy} introduce a new side-channel exploiting speculative decoding in LLMs, which is used as a latency optimization. They observe that input-dependent token count variations introduced by correct and incorrect speculative decoding can be used to fingerprint and leak private user inputs, or leak private or confidential data used for generating predictions. 
These variations in token counts are observed based on variations in packet sizes in streaming LLMs.

Song et al.~\cite{song2024earlybirdcatchesleak} introduced a timing side-channel attack exploiting the shared key-value (KV) cache and semantic cache that can be shared across multiple users on LLM serving platforms as a latency optimization. They demonstrated Prompt Stealing Attacks by incrementally searching system prompts or known documents in retrieval-augmented generation (RAG) systems. Moreover, their Peeping Neighbour Attack recovered semantically similar prompts from other users by exploiting the semantic cache. 

In contrast to these approaches, which rely on signals from performance optimizations which can be turned off, our method exploits timing signals from the autoregressive decoding process, which is inherent in all modern Transformer-based LLMs, making it significantly harder to mitigate. 
Moreover, unlike Song et al.~\cite{song2024earlybirdcatchesleak} who utilize timing signals originating from the prefill phase, our  work is grounded in the timing variations from the autoregressive decode-phase, which makes up the dominant part of the LLM generation time; thus our attack exploits stronger timing variations and signals compared to prior work~\cite{song2024earlybirdcatchesleak}.

\subsection{Timing Side-Channels in Machine Learning}
Several previous attempts use the timing as the observation channel to reveal the structure of DNNs. Naghibijouybari et al. \cite{10.1145/3243734.3243831} propose to use the co-location of attacker and victim in the resource space of Nvidia GPUs to interleave execution. The side channels are then exploited through memory allocation, program counters and time measurement to leak side-channel information. 

Wei et al. \cite{9153424} use the GPU context-switching penalty to reveal secret information about DNN networks, including layer architecture and hyperparameters. 
By turning off the Multi-Process Service, they force context switching between the victim and the attacker to achieve a much higher sampling rate. 
With this technique, they target the training stage since it takes a longer time and requires recurring execution of the same layer sequence. 
To adapt to the low sampling rate of the performance profiler, they launch more kernels inside the spy program to slow down the victim kernel due to the time-sliced scheduler. Then, operations and hyper-parameters can be inferred due to the differences in execution time. Additionally, they design voting models based on the Long Short-Term Memory model to identify different operations and hyperparameters.

In contrast, our work doesn't aim to recover the architecture or parameters of the machine learning model itself, but rather extract sensitive information inside the user input. Additionally, we don't use the microarchitectural hardware behaviour to observe the timing, but from a purely algorithmic perspective that exploits autoregression in LLMs.

\section{Conclusion}
\label{sec:conclusion}
This paper presents a novel timing side channel attack that leverages timing side channels to leak private attributes in user inputs by exploiting autoregressive decoding inherent in all modern-day LLMs. Our experiments illustrate a strong correlation between execution time, the number of generated tokens, and sensitive attributes in user inputs for LLMs dedicated to specialized tasks. These timing signals can leak the number of generated output tokens, allowing attackers to execute this attack in network scenarios. We evaluate this attack on machine translation and few-shot binary classification tasks. By analyzing timing and the lengths of input and output, we show an attacker can accurately leak the target language and classification results. To address this, we propose mitigations at various stages, including tokenization, prompt design, and system architecture. With this work, we highlight that developers need to consider variations in output tokens as a vector for information leakage while deploying LLM-based applications.

\newpage
\bibliographystyle{IEEEtran}
\bibliography{bibliography}

\appendix
\begin{appendices}
\newcommand{\thickhline}{\noalign{\hrule height 1pt}}
\begin{table*}
\caption{Translation performance (BLEU scores) for different models, M2M100, MBart50, Tower.}
\centering
\begin{tabular}{lcccccccccccc}
    \thickhline
     & Chinese & French & Hindi & Spanish & Arabic & Portuguese & Russian & German & Japanese & Korean & Italian & Dutch \\
     \hline
M2M100   & 19.0    & 44.7   & 26.6  & 25.4    & 13.1   & 45.6       & 28.1    & 35.4   & 28.5     & 21.9   & 26.8    & 25.0  \\
     \hline
MBart50  & 19.0    & 42.6   & 22.5  & 21.4    & 17.1   & 33.8       & 24.8    & 34.1   & 28.0     & 19.6   & 24.4    & 22.6  \\
     \hline
Tower    & 21.4    & 49.3   & -     & 27.2    & -      & 49.8       & 30.4    & 37.3   & -        & 24.4   & 29.8    & 27.3  \\
     \hline
\end{tabular}
\label{tab:translation_performance}
\end{table*}

\section{Validating Translation Model Performance}\label{app:translation_validation}
\cref{tab:translation_performance} shows translation performance of three models—M2M100, MBart50, and Tower—across various languages, measured using BLEU scores. Overall, all models achieve sufficiently high accuracy, with the Tower model generally outperforming the others, especially in languages like French, Portuguese, and Dutch. High scores in widely spoken languages reflect effective training and robust multilingual capabilities, while reasonable performance in less-resourced languages like Chinese and Korean highlights the models' ability to generalize. These models are thus well-suited for diverse multilingual translation tasks.
\vspace{1in}
~~ 
~~ 
~~ 
~~ 
~~ 
~~ 
~~ 
~~ 
~~ 
\vspace{2in}
\section{Additional Analysis for Translation Attack}
\label{app:translation_extra_analysis}
~~ 

\begin{figure}[h]
  \centering
  \includegraphics[width=\linewidth]{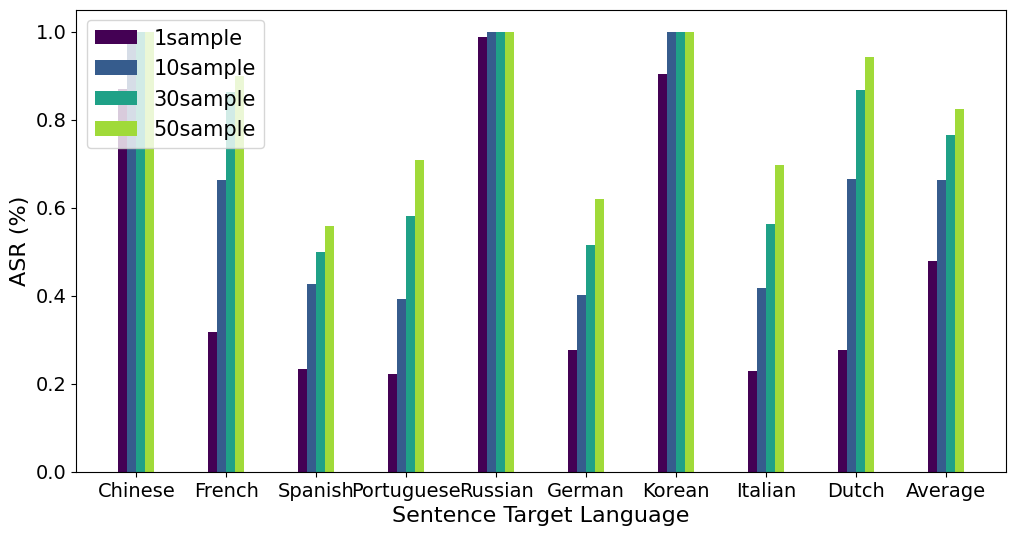}
  \caption{Attack success rate (ASR) for the translation attack on the Tower model as the number of sampled requests from the user increase from 1 to 10 to 30 to 50 samples (default = 50).}
  \label{fig:precision_tower_samples}
\end{figure}

\begin{figure}[h!]
  \centering
  \includegraphics[width=\linewidth]{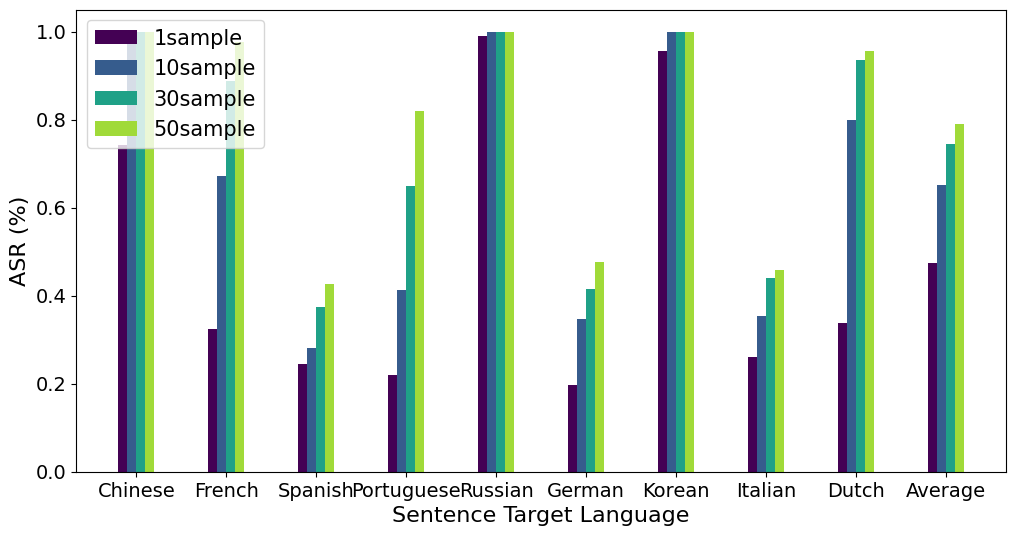}
  \caption{Attack success rate (ASR) for the translation attack on Tower model using EuroParl dataset as test set (retaining Flores as training set).}
  \label{fig:precision_tower_europarl}
\end{figure}

\vspace{2in}

\section{Translation Attack vs Source Language}\label{app:translation_src_languages}
~
\begin{table}[h]
\centering
\caption{ASR (\%) for Recovering Target Language as Source Languages Vary for Tower model}
\begin{tabular}{|c|c|c|c|c|}
\hline
\textbf{Source $\rightarrow$} & \textbf{English} & \textbf{French} & \textbf{Spanish} & \textbf{Russian} \\\hline
English & - & 100.0 & 100.0 & 100.0 \\\hline
Chinese & 100.0 & 100.0 & 100.0 & 100.0 \\\hline
French & 90.7 & - & 89.6 & 69.3 \\\hline
Spanish & 51.8 & 77.7 & - & 36.7 \\\hline
Portuguese & 72.8 & 76.2 & 87.7 & 76.7 \\\hline
Russian & 100.0 & 100.0 & 100.0 & - \\\hline
German & 63.1 & 92.7 & 86.2 & 42.9 \\\hline
Korean & 100.0 & 100.0 & 100.0 & 100.0 \\\hline
Italian & 73.4 & 85.5 & 79.3 & 79.3 \\\hline
Dutch & 97.0 & 95.9 & 97.3 & 94.4 \\\hline
\textbf{Average} & \textbf{83.2} & \textbf{92.0} & \textbf{93.3} & \textbf{77.7} \\\hline
\end{tabular}
\label{tab:tower_source_languages}
\end{table}

\begin{table}[h!]
\centering
\caption{ASR (\%) for Recovering Target Language as Source Languages Vary for M2M100 model}
\begin{tabular}{|c|c|c|c|c|}
\hline
\textbf{Source $\rightarrow$} & \textbf{English} & \textbf{French} & \textbf{Spanish} & \textbf{Russian} \\\hline
English & - & 97.5 & 100.0 & 99.8 \\\hline
Chinese & 100.0 & 100.0 & 100.0 & 100.0 \\\hline
French & 83.4 & - & 87.3 & 68.6 \\\hline
Spanish & 63.5 & 83.5 & - & 55.3 \\\hline
Portuguese & 75.3 & 67.2 & 65.2 & 62.9 \\\hline
Russian & 100.0 & 100.0 & 100.0 & - \\\hline
German & 44.0 & 68.9 & 60.6 & 59.1 \\\hline
Korean & 60.4 & 64.9 & 70.8 & 55.6 \\\hline
Italian & 38.7 & 48.7 & 37.3 & 36.4 \\\hline
Dutch & 45.4 & 54.2 & 49.5 & 67.0 \\\hline
Hindi & 100.0 & 100.0 & 100.0 & 100.0 \\\hline
Arabic & 100.0 & 100.0 & 100.0 & 100.0 \\\hline
Japanese & 98.5 & 91.3 & 94.5 & 98.6 \\\hline
\textbf{Average} & \textbf{75.8} & \textbf{81.3} & \textbf{80.4} & \textbf{75.3} \\\hline
\end{tabular}
\label{tab:m2m100_source_languages}
\end{table}

\vspace{2in}
\section{Ablation Study for Translation Attack} 
\label{app:translation_ablation}
~~
\begin{figure}[h]
  \centering
  \includegraphics[width=\linewidth]{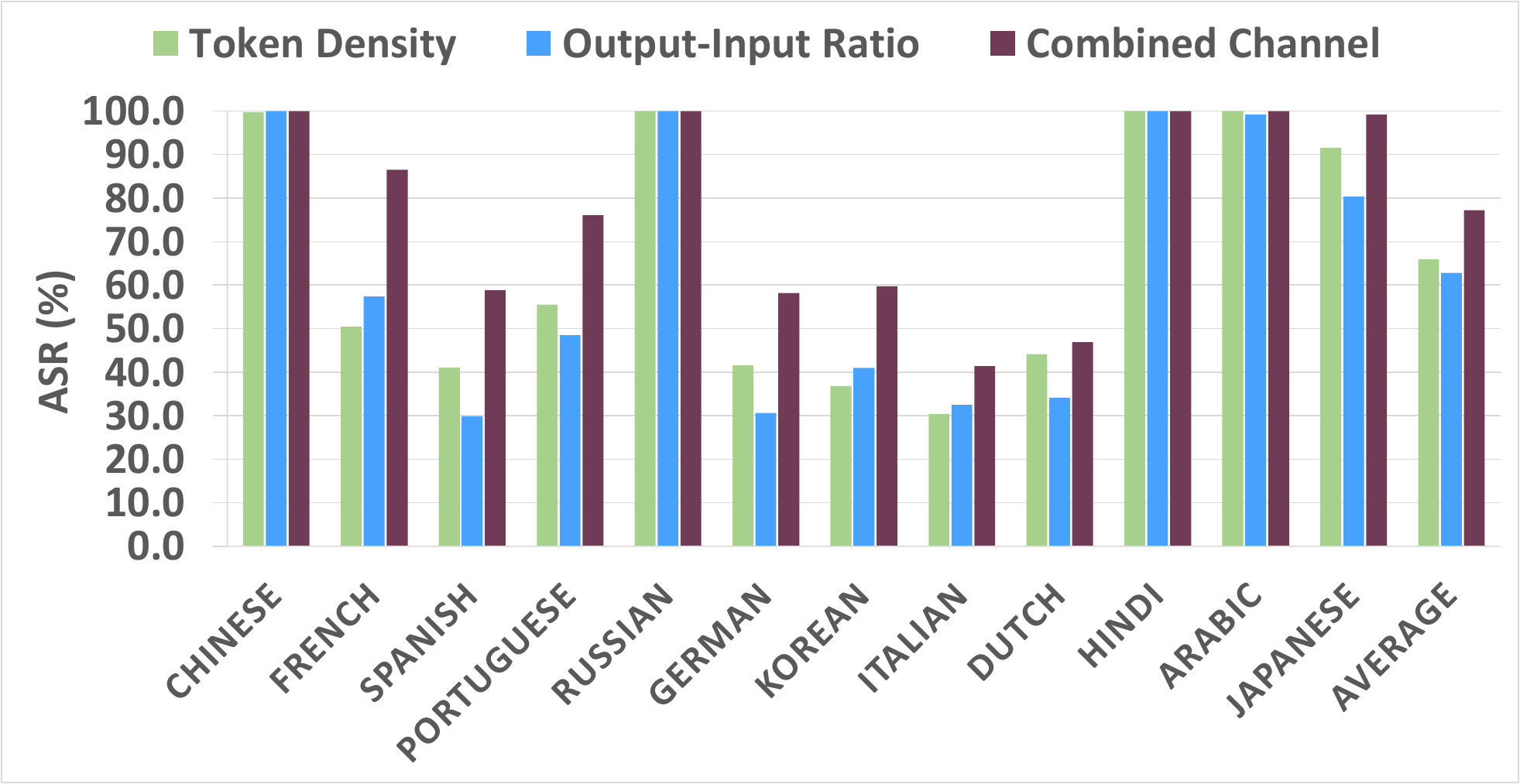}
  \caption{Ablation study for ASR (\%) using just token density or just output-input length ratio for M2M100 model, versus combined channel (default).}
  \label{fig:m2m100_ablation}
\end{figure}

\begin{figure}[h]
  \centering
  \includegraphics[width=\linewidth]{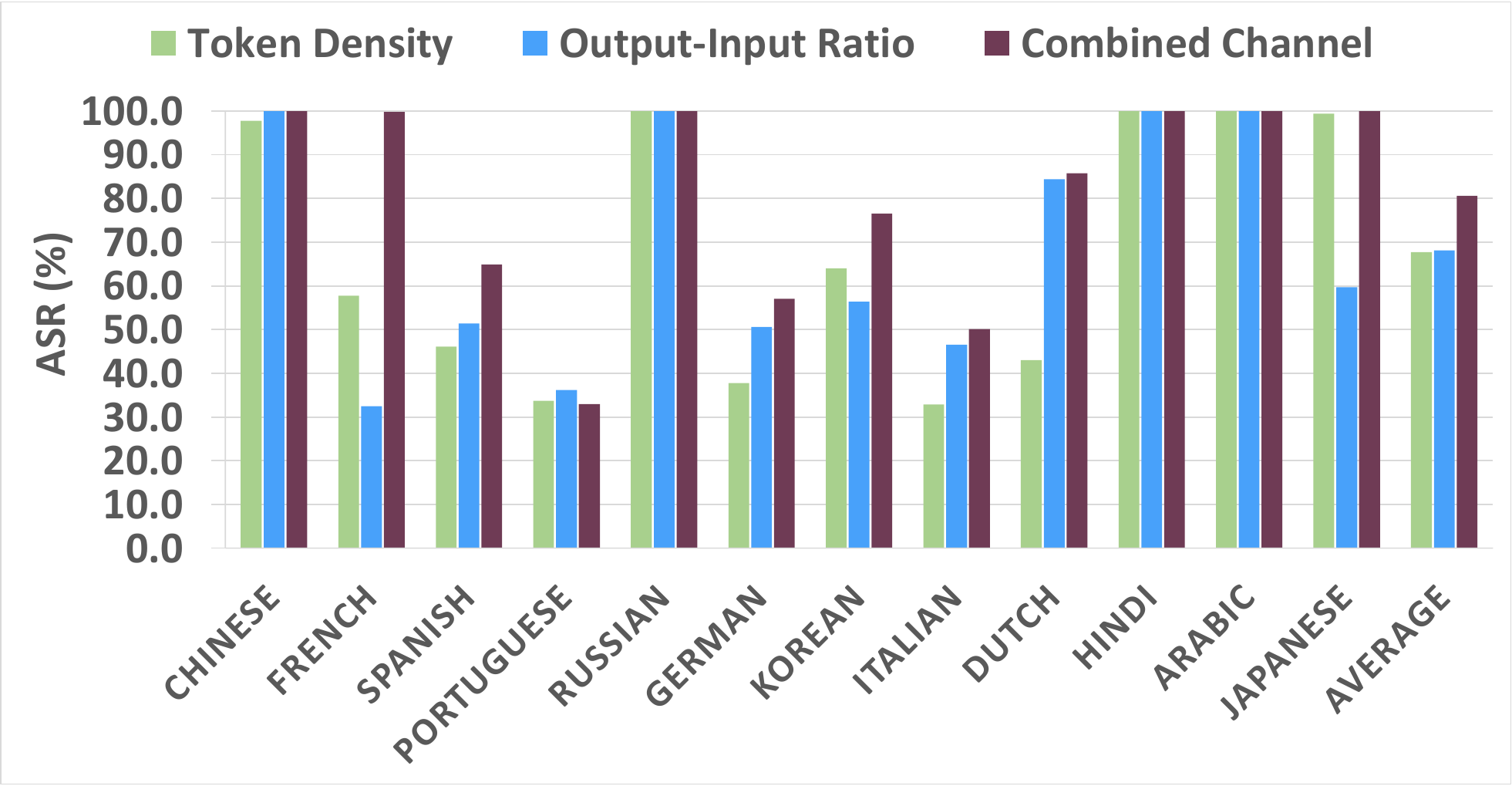}
  \caption{Ablation study for ASR (\%) with just token density or just output-input length ratio for MBart50, versus combined channel (default).}
  \label{fig:mbart50_ablation}
\end{figure}

\vspace{0.5in}
\section{Ablation Studies for Classification Attack}\label{app:abl_classification_attack}
~~
\begin{table}[h]
\caption{ASR (\%) as the temperature varies, for the augmenting 3-shot examples.}
\centering
\begin{tabular}{|c|c|c|c|c|c|c|}
\hline
\multirow{2}{*}{\textbf{Task \#}} & \multicolumn{2}{|c|}{\textbf{Temp=0}} & \multicolumn{2}{|c|}{\textbf{Temp=0.3}} & \multicolumn{2}{|c|}{\textbf{Temp=0.7}} \\  \cline{2-7}
& Gemma2-9B & GPT & Gemma2-9B & GPT & Gemma & GPT \\ \hline
227 & 83.5 & 91.0 & 81.5 & 88.3 & 77.0 & 89.3 \\ \hline
391 & 81.8 & 80.3 & 80.6 & 84.6 & 79.2 & 85.8 \\ \hline
392 & 78.8 & 82.3 & 78.7 & 85.0 & 77.3 & 86.7 \\ \hline
590 & 75.7 & 87.8 & 72.2 & 86.0 & 74.7 & 84.5 \\ \hline
879 & 94.7 & 93.3 & 97.3 & 91.7 & 95.6 & 92.5 \\ \hline
\textbf{Average} & \textbf{82.9} & \textbf{86.9} & \textbf{82.1} & \textbf{87.1} & \textbf{80.8} & \textbf{87.8} \\ \hline
\end{tabular}
\label{tab:asr_fewshot_temperature}
\end{table}

\begin{table}[h]
\caption{ASR (\%) only for results with only correct model predictions, with Augmenting 3-shot examples (difference with ASR across all model predictions in brackets).}
\centering
\begin{tabular}{|c|c|c|}
    \hline
    \textbf{Task \#}& \textbf{Gemma2-9B} & \textbf{GPT-4o} \\ \hline
    145 & 80.8 (+2.0) & 91.7 (+1.2) \\ \hline
    146 & 84.9 (+4.4) & 96.1 (+2.8) \\ \hline
    147 & 85.8 (+0.5) & 95.1 (+3.4) \\ \hline
    227 & 82.0 (-1.5) & 89.9 (-1.1) \\ \hline
    391 & 81.3 (-0.5) & 82.7 (+2.4) \\ \hline
    392 & 80.1 (+1.3) & 83.0 (+0.7) \\ \hline
    577 & 86.9 (+8.1) & 93.8 (+4.1) \\ \hline
    590 & 75.1 (-0.6) & 87.6 (-0.2) \\ \hline
    607 & 85.6 (+1.8) & 88.4 (-1.8) \\ \hline
    673 & 86.3 (+0.6) & 90.6 (+0.4) \\ \hline
    679 & 38.6 (-4.9) & 82.7 (+1.0) \\ \hline
    879 & 94.6 (-0.1) & 94.9 (+1.6) \\ \hline
    \textbf{Average} & \textbf{80.2 (+0.9)} & \textbf{89.7 (+1.2)} \\ \hline
\end{tabular}
\label{tab:asr_correct_incorrect}
\end{table}

\section{End-To-End Attack Profiling}\label{app:end_end_profiling}

\begin{figure}[h!]
  \centering
  \includegraphics[width=0.45\textwidth]{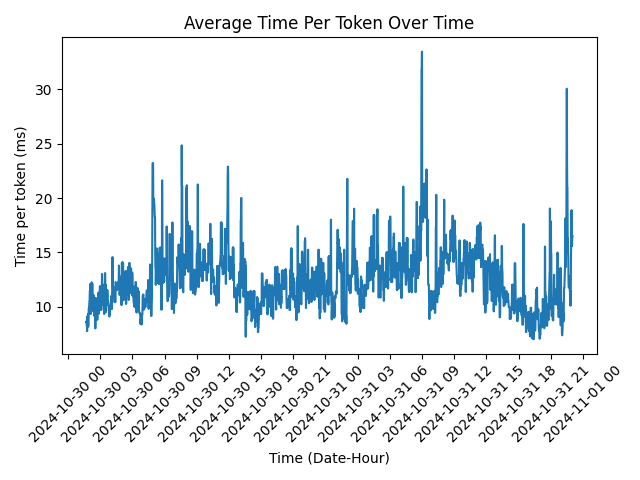}
  \caption{Time per Output Token (TPOT) Profiling for GPT-4o.}
  \label{fig:time_per_token}
\end{figure}

\begin{figure}[h!]
  \centering
  \includegraphics[width=0.45\textwidth]{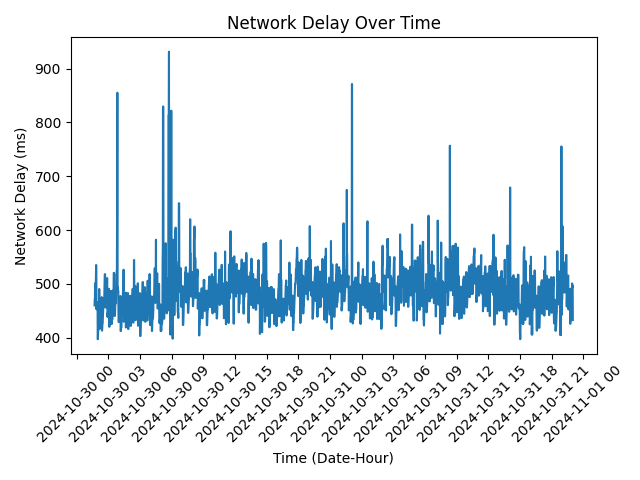}
  \caption{Network Delay Profiling for GPT-4o responses.}
  \label{fig:network_delay}
\end{figure}

\begin{table*}
\caption{Classification Tasks from Natural Instructions Dataset Used in Sections 5 and 6}
\label{tab:fewshottasks}
\centering
\begin{tabular}{|l|p{7cm}|p{7cm}|l|}
\hline
\textbf{Task \#} & \textbf{Task Definition} & \textbf{Sample User Input} & \textbf{Sample Output} \\
\hline
task145 & We would like you to classify each of the following sets of argument pairs (discussing Death Penalty) into either SIMILAR or NOT SIMILAR. A pair of arguments is considered SIMILAR if the arguments are about the same FACET (making the same argument), and is considered NOT SIMILAR if they do not have the same FACET. A FACET is a low-level issue that often reoccurs in many arguments in support of the author's stance or in attacking the other author's position. & Sent1: Yes there should be a death penalty but it should only be used in extreme circumstances like treason, mass murder, ordering murders from within prison, and killing someone in prison. Sent2: If a person just raped someone (this is just an example) they shouldn't be killed for it, but something like a mass murdering or serial killing should be dealt with by putting them to death. & Similar \\
\hline
task146 & We would like you to classify each of the following sets of argument pairs (discussing Gun Control) into either SIMILAR or NOT SIMILAR. A pair of arguments is considered SIMILAR if the arguments are about the same FACET (making the same argument), and is considered NOT SIMILAR if they do not have the same FACET. A FACET is a low-level issue that often reoccurs in many arguments in support of the author's stance or in attacking the other author's position. & Sent1: So now you're saying that just because somebody owns a gun, they're somehow more likely to start killing people than somebody who doesn't own a gun? Sent2: Sure you say guns don't kill people but people kill people, but the gun is only a tool or a means to that end. & Not similar \\
\hline
task147 & We would like you to classify each of the following sets of argument pairs (discussing Gay Marriage) into either SIMILAR or NOT SIMILAR. A pair of arguments is considered SIMILAR if the arguments are about the same FACET (making the same argument), and is considered NOT SIMILAR if they do not have the same FACET. A FACET is a low-level issue that often reoccurs in many arguments in support of the author's stance or in attacking the other author's position. & Sent1: I also think that most Americans are truly uninformed when it comes to the issue of same sex marriage. Sent2: I think the real issue is not "should gays be allowed to marry." & Not similar \\
\hline
task227 & In this task, you are given an ambiguous question/query (which can be answered in more than one way) and a clarification statement to understand the query more precisely. Your task is to classify whether the given clarification accurately clarifies the given query or not and based on that provide 'Yes' or 'No'. & Query: Tell me about Barbados. Clarification: do you want to file your income tax online & No \\
\hline
task391 & In this task, you will be given two sentences separated by ", so". You should decide whether the first sentence can be the cause of the second sentence. If you can see a possible causation, answer with "plausible", otherwise answer with "not plausible". & The terrorist set off the bomb, so they were able to kill him & plausible \\
\hline
task392 & In this task, you will be given two sentences separated by "because". You should decide whether the first sentence can be the result of the second sentence. If you can see a possible cause and effect relationship, answer with "plausible", otherwise answer with "not plausible". & I learned how to play the board game because I love to play it & plausible \\
\hline
task577 & In this task, you are given a dialogue between a user and an assistant, where users and assistants converse about geographic topics like geopolitical entities and locations. The task here is to find if the dialogue is by the user or assistant. Classify your answers into user and assistant. & Hi. Could you tell me about the culture and society of Pakistan? & user \\
\hline
task590 & In this task, you are given an Amazon food product review and its summary. Your task is to generate "True" if the given review and its summary match, otherwise generate "False". & This is great dog food, my dog has severe allergies and this brand is the only one that we can feed him. Summary: Great dog food & True \\
\hline
task607 & In this task, you are given a text from a social media post. Your task is to classify the given post into two categories: 1) yes if the given post is intentionally offensive, 2) no, otherwise. Also, generate label 'no' for offensive statements that appear to be unintentional or non-offensive statements that are underhandedly offensive. Warning: the examples and instances may contain offensive language. & RT @rihanna: I ain't got no type...bad bitches is the only thing that I like & No \\
\hline
task673 & Read the given query and classify it as a 'Good' or 'Bad' query depending on how well the query is formed, 'Bad' being the expected output for a not so well-formed query and 'Good' being the expected output for a well-formed query. A query may be wrong based on common sense or general facts, but if it is well-formed, you should answer with Good. & What is the torqur for the intake manifold for a 1999 Pontiac Grand Am 3.4 l ? & Good \\
\hline
task679 & Classify the given comment into 'Hope Speech' or 'Not Hope Speech'. A hope speech is defined as content that is encouraging, positive or supportive content and talks about equality, diversity or inclusion & Genuine question...why do many black people who complain about being oppressed (rightly so) not support other communities such as Palestinians & Not Hope Speech \\
\hline
task879 & You are given a sentence from a conversation between a human and a virtual assistant. Your task is to identify whether the sentence is a question or not. Answer with Yes or No. & Which city should I search? & Yes \\
\hline
\end{tabular}
\end{table*}

\end{appendices}

\end{document}